\documentclass[review,authoryear,12pt]{elsarticle}
\usepackage[english]{babel}
\usepackage{graphicx}
\usepackage{subfigure}
\usepackage{amssymb}
\usepackage{amsthm}
\usepackage{lineno}
\usepackage{multirow}
\usepackage{paralist}
\usepackage{tabularx}
\graphicspath{{images/}}
\journal{Pattern Recognition Letters}

\begin{document}

\begin{frontmatter}

\title{Invariant texture analysis through Local Binary Patterns}

\author[nava]{Rodrigo Nava}
\address[nava]{Posgrado en Ciencia e Ingenier\'ia de la Computaci\'on, Universidad Nacional Aut\'onoma de M\'exico, Mexico City, Mexico}
\ead{urielrnv@uxmcc2.iimas.unam.mx}

\author[cristobal]{Gabriel Crist\'obal}
\address[cristobal]{Instituto de \'Optica, Serrano 121, 28006 Madrid, Spain}
\ead{gabriel@optica.csic.es}

\author[escalante]{Boris Escalante-Ram\'irez}
\address[escalante]{Facultad de Ingenier\'ia, Universidad Nacional Aut\'onoma de M\'exico, Mexico City, Mexico}
\ead{boris@servidor.unam.mx}

%-------------------------------------------------------------------------
\begin{abstract}
In many image processing applications, such as segmentation and classification, the selection of robust features descriptors is crucial to improve the discrimination capabilities in real world scenarios. In particular, it is well known that image textures constitute power visual cues for feature extraction and classification. In the past few years the local binary pattern (LBP) approach, a texture descriptor method proposed by Ojala et al., has gained increased acceptance due to its computational simplicity and more importantly for encoding a powerful signature for describing textures. However, the original algorithm presents some limitations such as noise sensitivity and its lack of rotational invariance which have led to many proposals or extensions in order to overcome such limitations. In this paper we performed a quantitative study of the Ojala's original LBP proposal together with other recently proposed LBP extensions in the presence of rotational, illumination and noisy changes. In the experiments we have considered two different databases: Brodatz and CUReT for different sizes of LBP masks. Experimental results demonstrated the effectiveness and robustness of the described texture descriptors for images that are subjected to geometric or radiometric changes.
\end{abstract}

%-------------------------------------------------------------------------

\end{frontmatter}

\linenumbers

%-------------------------------------------------------------------------
\section{Introduction~\label{sec:intro}}
Texture is the term used to characterize object surfaces and is used for pattern identification. It has been studied in the fields of visual perception and computer vision. Although it is a feature often used to characterized objects, it has been difficult to establish an appropriate definition. Since a texture is quite varied and can exhibit a large number of properties, many vision researchers have given definitions frequently in the context of different applications areas, \cite{TUC1998}. However, from a mathematical point of view, it is usual to analyze textures as intensity variations from regularity --when textures simply contain periodic patterns-- to randomness --where textures look like unstructured noise.

There are many ways to classify textures. \cite{HARALICK1979} proposed two different approaches: the statistical and the structural methods. The first one considers textures as the arrangement of spatial distribution of gray values in images. Inside of this group one can highlight the features extracted from the co-occurrence matrix, \cite{DAVIS1979}. Structural methods are based on considering that textures are composed by primitives called ``textons''. In this direction, \cite{WANG1990} introduced a model where textures can be characterized by its texture spectrum, a set of essential small units. 

A more detailed texture classification was later proposed by \cite{PAGET2008}, the approaches may be divided into the following categories: \begin{inparaenum}[\itshape i\upshape)]
\item \textbf{statistical methods:} a set of features is used to represent textures. The basic assumption is that the intensity variations are more or less constant within a texture region and takes a greater value outside its boundary. Statistical measures analyze spatial distribution of pixels using features extracted from the first and second-order histogram statistics \cite{GUO2010b}.
\item \textbf{spectral methods:} these methods collect a distribution of filter responses as input to further classification or segmentation. In particular, Gabor filters have proven to be powerful and precise for describing texture patterns, \cite{NAVA2011}. Novel approaches seem to lead towards an improvement of Gabor filters by using local binary patterns as a complementary tool to extract texture features as in \cite{MA2007, NGUYEN2009, Zhang2005}. Many algorithms in this category are focusing on face recognition, \cite{HUANG2011}. In addition, a recent analysis of rotational invariant texture features appears in \cite{Estudillo2011}. 
\item \textbf{structural methods:} some textures can be viewed as two dimensional patterns consisting of a set of primitives which are arranged according to a certain placement rules.
\item \textbf{stochastic methods:} textures are assumed to be the realization of a stochastic process. The parameter estimation associated with the process is quite complicated although there are good approaches in the literature, e.g., \cite{SEETHARAMAN2009} use a Bayesian approach as a texture descriptor.
\end{inparaenum}

Texture analysis through a LBP operator can be considered as a combination of both statistical and structural methods. Therefore, it can be expected a good LBP performance in a wide variety of texture identification scenarios. LBP constitutes an image operator that transforms an image into an array of integer labels that encode the pixel-wise information of the texture images. These labels can be represented as a histogram that can be interpreted as the fingerprint of the analyzed object. In fact, the LBP approach belongs to a group of non-local parametric transformations that is distinguished by the use of ordering information among data, rather than the data values themselves. Non-parametric local transformations are local image transformations that rely on the relative ordering of the intensity values.

Similarly to the \cite{OJALA1994} work, \cite{ZABIH1994} proposed two alternative non-parametric local transforms. The first transform called rank transform (RT) is defined as the number of pixels in a local square region whose values are lesser that the value of a central pixel. The second non-parametric local transform named census transform (CT) maps the local square neighborhood into a bit string representing the set of neighbor pixels whose intensities are lesser than a central pixel value. Both RT and CT depend solely on a set of pixel comparisons. The first limitation of these kind of methods is that the amount of information associated to a pixel is not very large which induces noise sensitivity. Another limitation is that the local measures rely heavily upon the intensity of a central pixel. Nevertheless, the last drawback is not an issue by doing comparisons using local means or median values instead of central pixel intensities \cite{ZABIH1994}.

After the initial LBP proposal, many modifications and improvements have emerged in the literature, most of them are related to face analysis where it is assumed that input faces are registered. For this reason, many modifications are not invariant to rotational transforms. For a thorough description of LBP operators see two recent surveys and a book monograph: \cite{NANNI2011, HUANG2011, PIETIKAINEN2011}.

This paper is organized as follows. Section \ref{sec:overview} presents an overview of LBP methods as well as the most significant extensions that have been proposed to obtain rotational invariance. In Section \ref{sec:experiments} an exhaustive evaluation of different LBP approaches previously mentioned as well as a comparative analysis of LBP rotational invariant proposals is shown. This study includes a few tests with noise and illumination changes using a rotated version of the Brodatz database, \cite{BRODATZ1966} and the CUReT texture database, \cite{DANA1999}. Finally our work is summarized in Section \ref{sec:conclusions}.

%-------------------------------------------------------------------------
\section{Local Binary Pattern Overview~\label{sec:overview}}
The very first approach to LBP was given in \cite{OJALA1994}. Ojala proposed a two-level version of the original method of \cite{WANG1990}. Ojala claimed that this refinement provides a robust way for describing local texture patterns. However, very recently \cite{TAN2010b} have revisited the original approach and demonstrated that a generalization of LBP called local ternary patterns (LTP) is more discriminant and less sensitive to noise for texture analysis. 

\begin{figure}[htbp]
    \centering
    \subfigure[]{\label{fig:fig0_a}\includegraphics[width=0.24\textwidth]{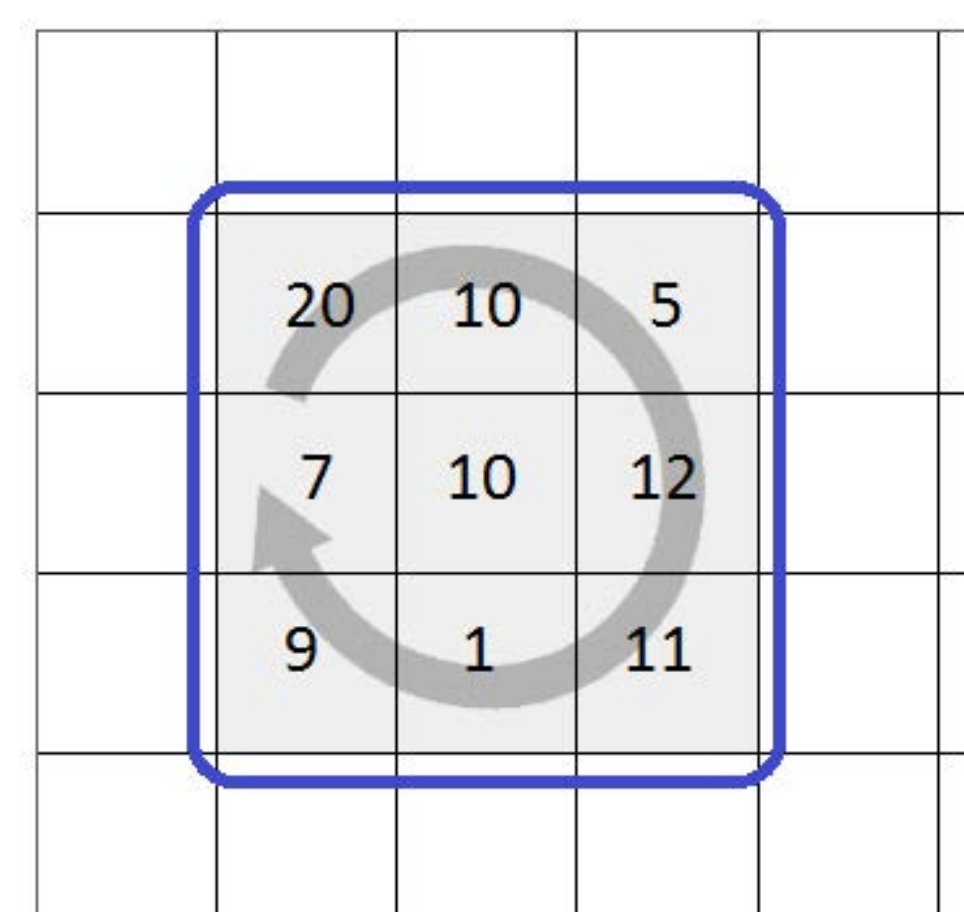}}
    \subfigure[]{\label{fig:fig0_b}\includegraphics[width=0.24\textwidth]{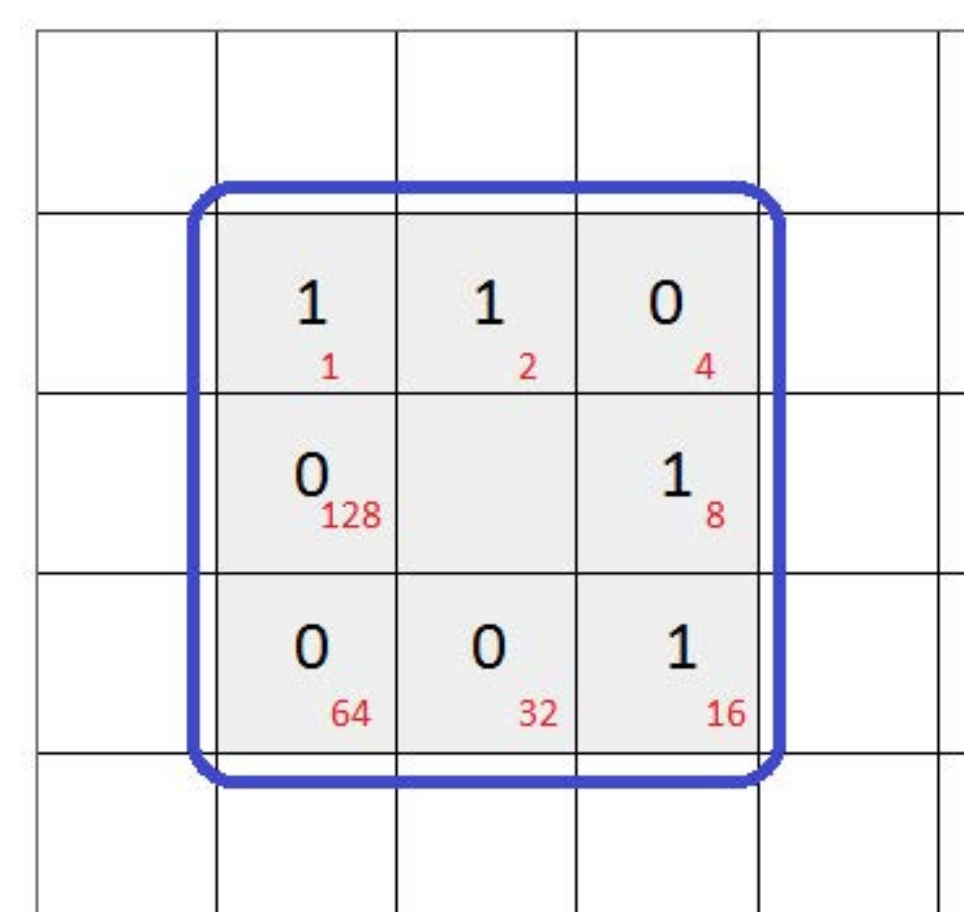}}
    \subfigure[]{\label{fig:fig0_c}\includegraphics[width=0.24\textwidth]{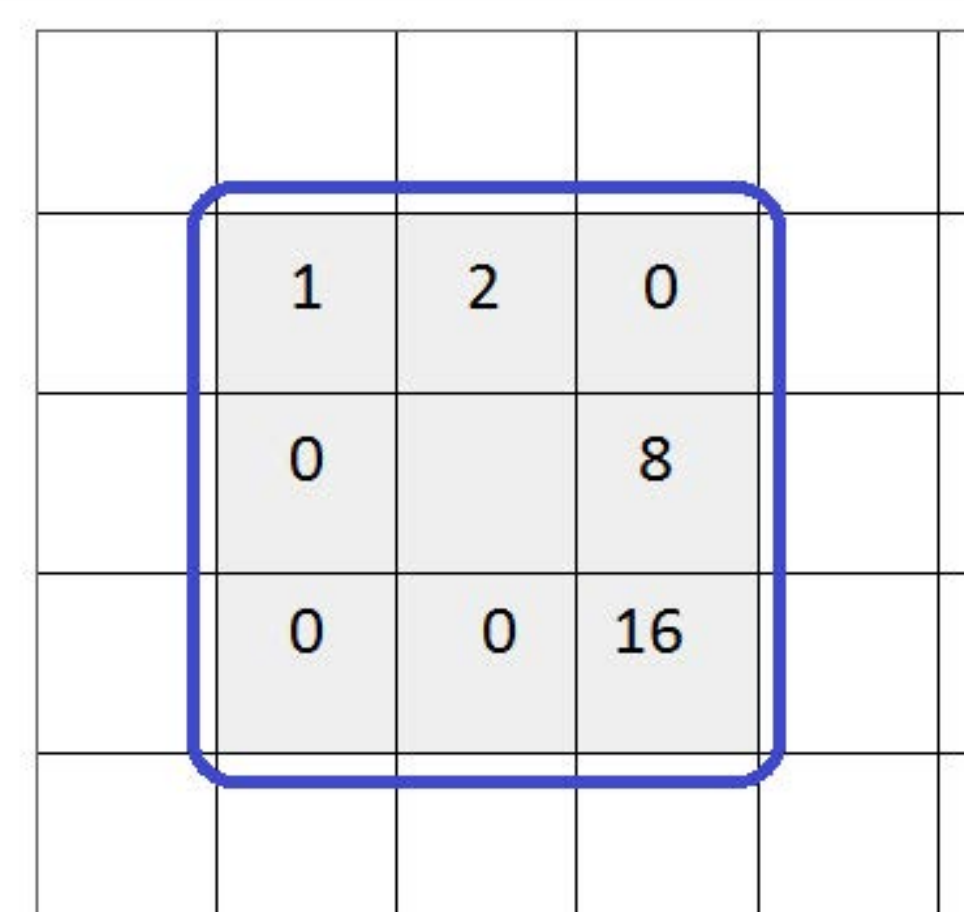}}
    \subfigure[]{\label{fig:fig0_d}\includegraphics[width=0.24\textwidth]{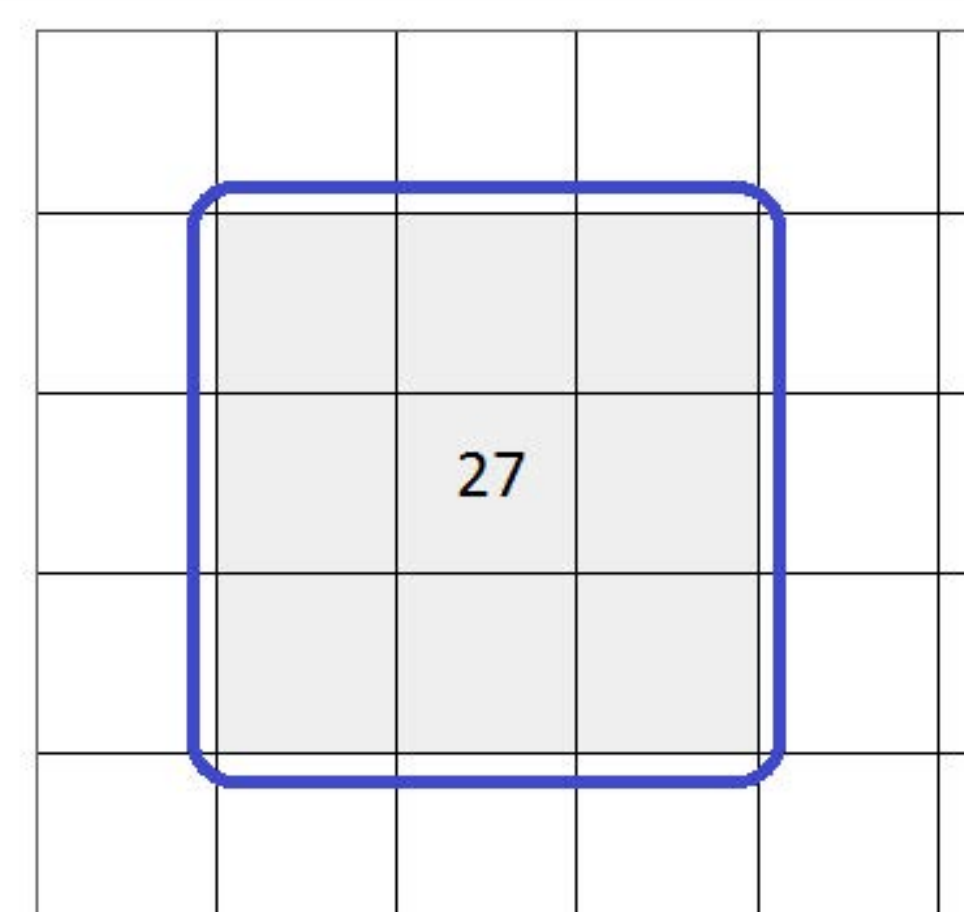}}
    \caption{Based on a square mask of $3\times 3$ the LBP algorithm computes the label by comparisons between central pixels and their surrounding neighbors.}
    \label{fig:fig0} 
\end{figure}

The simple LBP uses a $3\times 3$ square mask called ``texture spectrum'' that represents the neighborhood around a central pixel, (see Fig.~\ref{fig:fig0_a}). The values of the neighbor pixels within the square mask are thresholded by the value of their central pixel. Pixel values under the threshold are labeled with ``$0$'' otherwise they are labeled with ``$1$'', Fig.~\ref{fig:fig0_b}. The labeled pixels are multiplied by a weight function according with their positions, Fig.~\ref{fig:fig0_c}. Finally, the values of the eight pixels are summed to obtain a label for this neighborhood, Fig.~\ref{fig:fig0_d}. This method produced $2^8$ possible labels. After this process is completed for the whole image, a label or LBP histogram is computed so that can be interpreted as a fingerprint of the analyzed object. Although this method provides information about local spatial structures, it is not invariant to rotational changes and does not include contrast information which has been demonstrated to be crucial to improve the discrimination of some textures.

The classic LBP operator was later generalized by \cite{OJALA2002}. Such generalization can be obtained using a circular neighborhood denoted by $\left(P,R\right)$ where $P$ represents the number of sampling points and $R$ represents the radius of the neighborhood. The sampling point coordinates $\left(x_{p},y_{p}\right)$ are calculated using the formula $\left(x_{c} + R\cos\left(\frac{2\pi p}{P}\right),y_{c} - R\sin\left(\frac{2\pi p}{P}\right)\right)$. When sampling coordinates do not fall at integer positions, the intensity value is bilinearly interpolated. This implementation is called interpolated LBP ($LBP_{P,R}$).

$LBP_{P,R}$ is defined as an ordered set of binary comparisons of pixel intensities between a central pixel and its surrounding neighbors as follows:
\begin{equation}
    LBP_{P,R}\left(g_{c}\right) = \sum_{p=0}^{P-1}{s\left(g_{p} - g_{c}\right)2^{p}}
\label{eq:eq1}
\end{equation}
where $g_{c}$ is the intensity value of the central pixel at $\left(x_{c}, y_{c}\right)$ coordinates and $\left\{g_{p} | p = 0, \ldots, P-1\right\}$ is the intensity value of the $p$-neighbor. The thresholding function $s\left(x\right)$ is defined as:
\begin{equation}
   s\left(x\right) = \left\{
  \begin{array}{lc}
    1 & \mbox{if $x \geq 0$}\\
    0 & \mbox{if $x < 0$}\\
  \end{array} \right.
\label{eq:eq2}
\end{equation}

Eq.~(\ref{eq:eq1}) represents a texture unit composed of $P+1$ elements (central pixel included). In total, there are $2^{P}$ possible texture units describing spatial patterns in a neighborhood of $P$ points. $LBP_{P,R}$ achieves invariance against any monotonic transformation by considering the sign of the differences in $s\left(g_{p} - g_{c}\right)$, which effectively corresponds to binary thresholding of the local neighborhood.

$LBP_{P,R}$ is defined on a circular neighborhood allowing to change the number of neighbors and the radius size. However, increasing the number of neighbors increases the information redundancy and the computational cost, which not always resulting in a more discriminant LBP label. In this direction, \cite{LIAO2007b} defined the elongated LBP (ELBP) based on an anisotropic neighborhood allowing to improve texture discrimination by implementing multi-orientation analysis. 

\begin{figure}[htbp]
    \centering
    \subfigure[]{\label{fig:fig1_a}\includegraphics[width=0.19\textwidth]{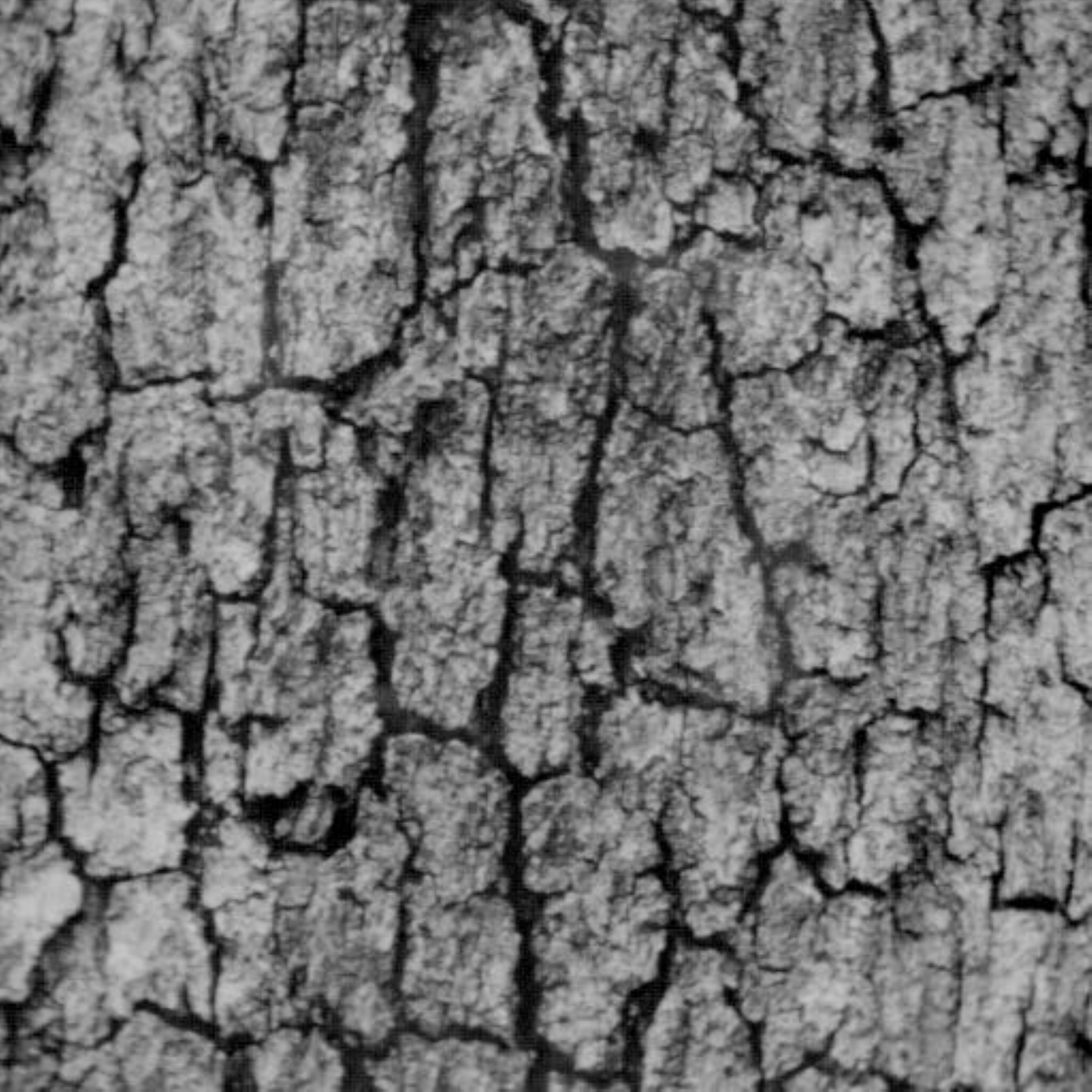}}
    \subfigure[]{\label{fig:fig1_b}\includegraphics[width=0.19\textwidth]{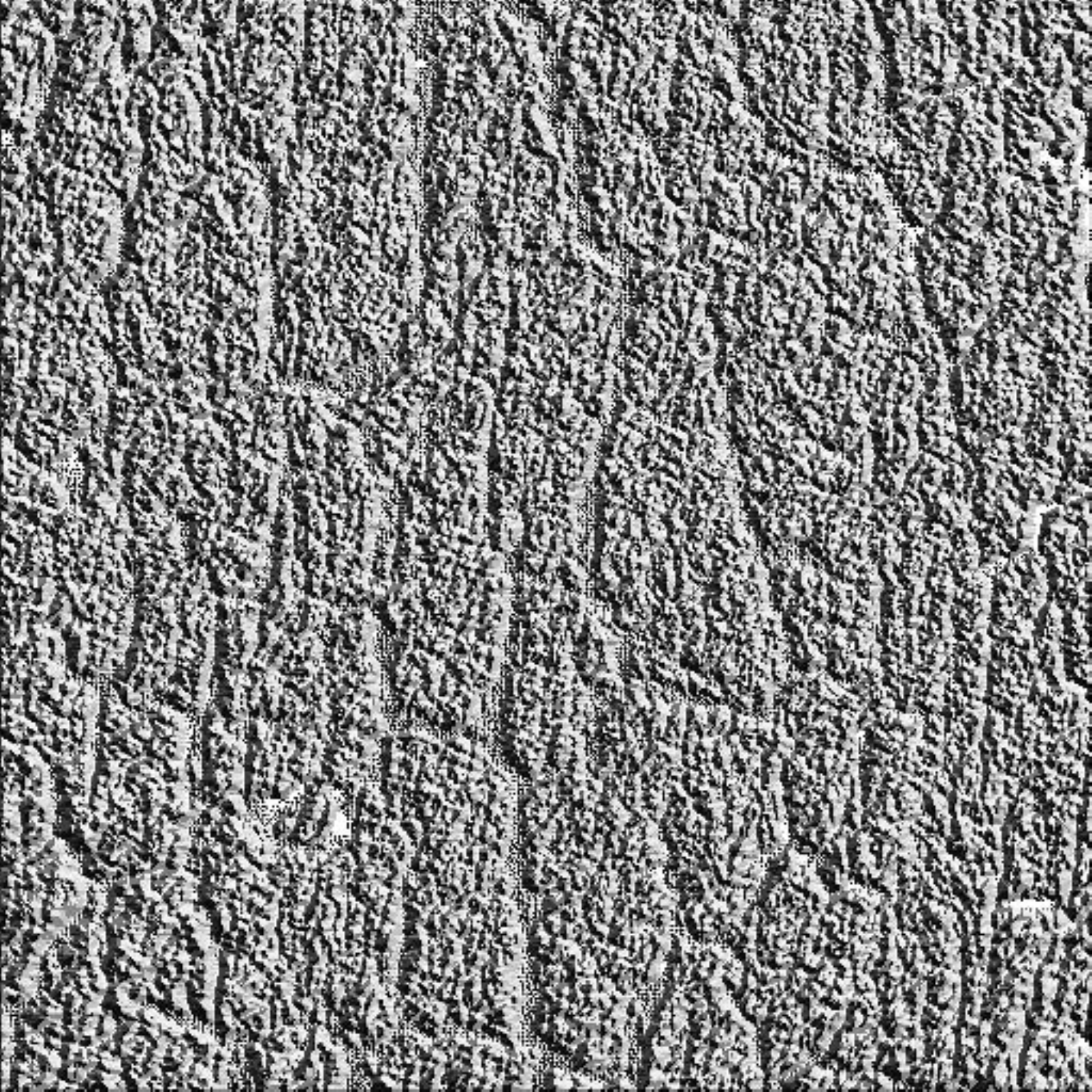}}
    \subfigure[]{\label{fig:fig1_e}\includegraphics[width=0.19\textwidth]{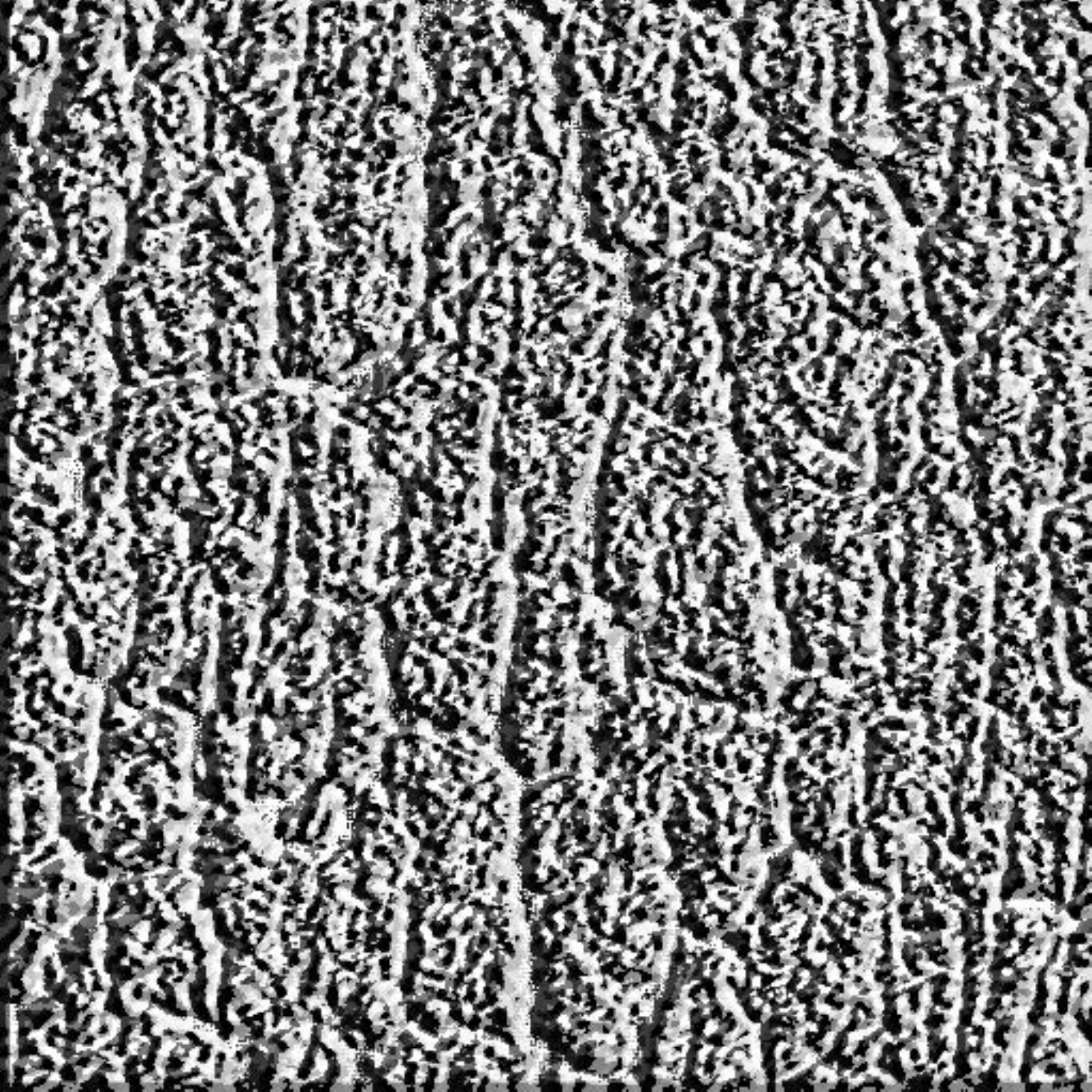}}
    \subfigure[]{\label{fig:fig1_f}\includegraphics[width=0.19\textwidth]{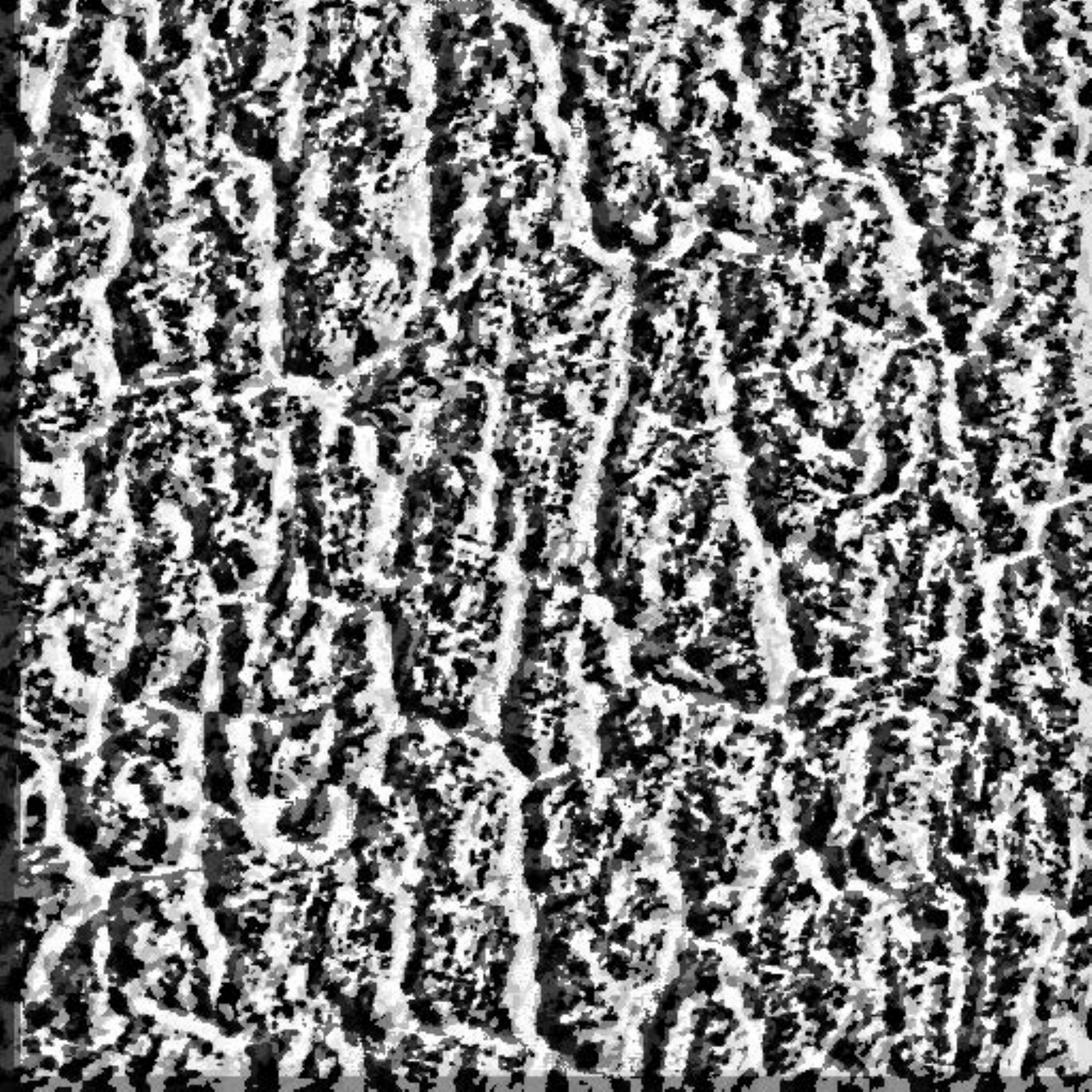}}
    \subfigure[]{\label{fig:fig1_g}\includegraphics[width=0.19\textwidth]{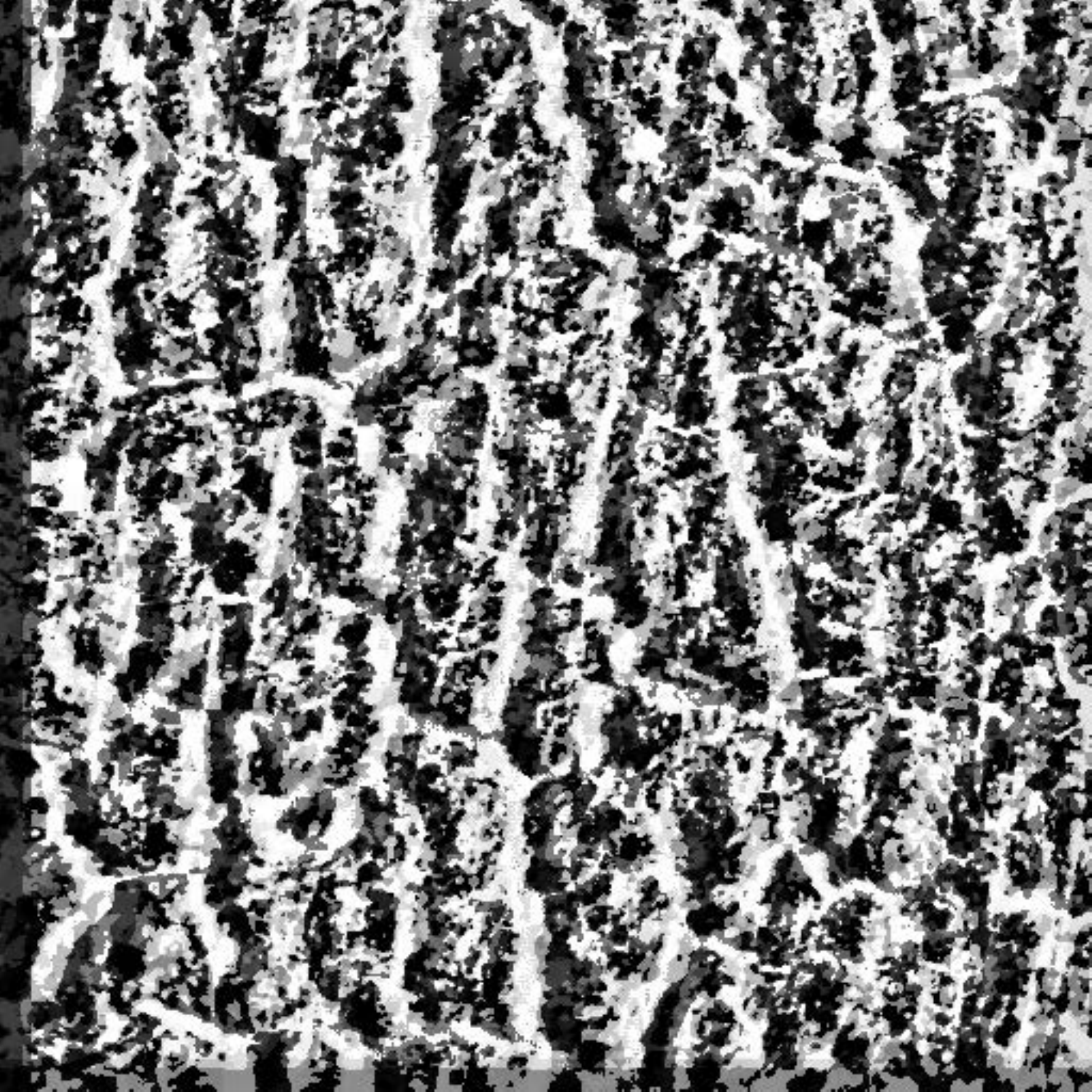}}
    \caption{LBP depends solely on the set of ordered comparisons between a central pixel and its surrounding neighbors. Specifically, $LPB_{P,R}$ allows to analyze neighborhoods with different sizes and number of neighbors. Nonetheless, the larger neighborhood size the coarse LBP image become. LBP images from Fig.~\ref{fig:fig1_b} to Fig.~\ref{fig:fig1_e} are texture processed by the LBP operator with different values of $P$ and $R$. \ref{fig:fig1_a} Bark texture (D12). \ref{fig:fig1_b} $P=8, R=1$. \ref{fig:fig1_e} $P=8, R=5$. \ref{fig:fig1_f} $P=8, R=10$. \ref{fig:fig1_g} $P=1, R= 15$. }
    \label{fig:fig1} 
\end{figure}

In relation with the size of the radius, \cite{LIAO2007a} proposed a representation called multi-scale block LBP (MBLBP). The computation is done based on averaging values of block subregions instead of individual pixels. In this paper will further analyze the influence of the neighborhood size, (see Fig.~\ref{fig:fig1}).

Under a rotational transform the values of $\left\{g_{p} | p = 0, \ldots, P-1\right\}$ will move along a circular path around a central pixel $g_{c}$ resulting into different labels. \cite{PIETIKAINEN2000} proposed a modification called rotational invariant LBP ($LBP^{min}_{P,R}$) to remove the rotational effects by labeling each rotation with an identifier as follows:
\begin{equation}
    LBP^{min}_{P,R}\left(g_{c}\right) = min\left\{ROR\left(LBP_{P,R}\left(g_{c}\right), i\right) | i = 0, \ldots, P-1\right\}
\label{eq:eq3} 
\end{equation}
where $ROR\left(x, i\right)$ performs a circular bitwise right shift operation $i$ times. 

The main idea is to rotate the $P$ neighbors to find the minimum value that the neighbor chain may represents. This approach identifies $36$ different values when using $P=8$. Nevertheless, Eq.~(\ref{eq:eq3}) achieves invariance only for a discrete digital domain because only for $90^{\circ}$ perfect rotational invariance can be attained.

\cite{OJALA2002} observed that over $90\%$ of LBPs entail fundamental properties of textures that can be described with very few spatial transitions. He introduced a uniformity measure $U\left(LBP_{P,R}\left(g_{c}\right)\right)$ which corresponds to the number of spatial transitions in the pattern as follows:
\small
\begin{equation}
    U\left(LBP_{P,R}\left(g_{c}\right)\right) = |s\left(g_{p-1} - g_{c}\right) - s\left(g_{0} - g_{c}\right)| + \sum_{p = 1}^{P-1}{|s\left(g_{p} - g_{c}\right) - s\left(g_{p} - g_{c}\right)|}
\label{eq:eq4}
\end{equation}
\normalsize
in this way the so-called uniform LBP ($LBP^{uni}_{P,R}$) can be obtained as:
\begin{equation}
   LBP^{uni}_{P,R}\left(g_{c}\right) = \left\{
  \begin{array}{l c}
    \sum_{p = 0}^{P-1}{s\left(g_{p} - g_{c}\right)} & \mbox{if $U\left(LBP_{P,R}\left(g_{c}\right)\right) \leq 2$}\\
    P+1 & \mbox{otherwise}\\
  \end{array} \right.
\label{eq:eq5}
\end{equation}

Eq.~(\ref{eq:eq5}) represents a gray-scale and rotational invariant texture descriptor that assigns a unique label to each patterns where the number of spatial transitions is at most two. These labels corresponding to the number of ``$1$'' in the pattern chain while the rest of the non-uniform patters (where the number of spatial transitions is greater than two) are grouped under the label $P + 1$. The discrete histogram of the uniform patterns obtained has been shown to be a very powerful feature for characterizing textures,~\cite{ZHOU2008}.

%-------------------------------------------------------------------------
\subsection{Modifications of rotational invariant LBP ~\label{sec:mod}}
In Eq.~(\ref{eq:eq5}) all patterns are divided according with the number of spatial transitions resulting in $P + 1$ sets of uniform patterns. One disadvantage is that patterns with more that two transitions are grouped into a unique label which leads to loss of discrimination. The reason for that due to the fact there not exists a general pattern that will be able to describe all textures. Every pattern is well suited for describing just a certain texture. \cite{MA2011} proposed the number LBP ($LBP^{num}_{P,R}$) as an extension of the $LBP^{uni}_{P,R}$ by dividing the non-uniform patterns into groups based on the number of ``$1$'' bits or ``$0$'' bits as follows:

\scriptsize
\begin{equation}
   LBP^{num}_{P,R}\left(g_{c}\right) = \left\{
  \begin{array}{l l} 
    \sum_{p = 0}^{P-1}{s\left(g_{p} - g_{c}\right)} & \mbox{if $\begin{array}{l} U\left(LBP_{P,R}\left(g_{c}\right)\right) \leq 2 \end{array}$}\\
    Num_{1}\left\{LBP_{P,R}\left(g_{c}\right)\right\} & \mbox{if $\begin{array}{l} U\left(LBP_{P,R}\right) > 2 \mbox{\hspace{0.35cm} and} \\
    Num_{1}\left\{LBP_{P,R}\left(g_{c}\right)\right\} \geq Num_{0}\left\{LBP_{P,R}\left(g_{c}\right)\right\}
    \end{array}$}\\
    Num_{0}\left\{LBP_{P,R}\left(g_{c}\right)\right\} & \mbox{if $\begin{array}{l} U\left(LBP_{P,R}\right) > 2 \mbox{\hspace{0.35cm} and}\\ 
    Num_{1}\left\{LBP_{P,R}\left(g_{c}\right)\right\} < Num_{0}\left\{LBP_{P,R}\left(g_{c}\right)\right\}
    \end{array}$}
\end{array} \right.
\label{eq:eq6}
\end{equation}
\normalsize
where $Num_{1}\left\{\bullet\right\}$ is the number if ``$1$'' and $Num_{0}\left\{\bullet\right\}$ is the number of ``$0$'' in the non-uniform pattern. 

Along the same direction, \cite{ZHOU2008} proposed an extension by dividing the non-uniform patterns according to their structural properties and merging them on the basis of their degree of similarity so that the final histogram reflects texture information more efficiently because it may represent the stochastic components of textures. Moreover,~\cite{MA2011} outperforms Zhou's results. 

\cite{LIU2011} stated that the probability of a central pixel depends only on its neighbors. In this way the neighbor intensity LBP ($LBP^{ni}_{P, R}$) can be defined by replacing the central pixel value with the average of its neighbors as follows: 
\begin{equation}
    LBP^{ni}_{P, R}\left(g_{c}\right) = \sum_{p=0}^{P-1}{s\left(g_{p}-\mu\right)}2^{p}
\label{eq:eq7}
\end{equation}
where
\begin{equation}
    \mu = \frac{1}{p}\sum_{p= 0}^{p-1}{g_{p}}
\label{eq:eq8}
\end{equation}

The presence of noise in images can seriously impair the texture extraction performance of the LBP operator. In this paper we propose to use the median operator in order to reduce noise effects. Such proposal replace the central pixel value with the median of itself and the $P$ neighbors as follows:
\begin{equation}
      LBP^{med}_{P, R}\left(g_{c}\right) = \sum_{p=0}^{p-1}{s\left(g_{p}-\widetilde{g}\right)}
\label{eq:eq9}
\end{equation}
where $\widetilde{g}$ represents the median of the $p$ neighbors and the central pixel, \cite{ZABIH1994}. This LBP modification is still invariant to rotation but less sensitive to noise. It is also invariant to monotonic illumination changes.

%-------------------------------------------------------------------------
\subsection{Other LBP extensions oriented to face analysis}
In the last few years, face image analysis has been one of the most active research areas where LBP has been exploited because its effectiveness in deal with various challenging task of face analysis. Since most of the face detection or segmentation algorithms include a normalization step --which means that faces are registered-- as a preprocessing step and therefore the LBP methods do not need to present affine invariant features. 

\cite{FU2008} addressed the problem of noise sensitivity by considering that in most cases central pixels provide more information than their neighbor counterparts, so they assigned to the central pixels a bigger weight. In the case of images degraded by white noise Fu and Wei considered a modified version of Eq.~(\ref{eq:eq2}) as:
\begin{equation}
   s\left(x\right) = \left\{
  \begin{array}{lc}
    1 & |x|\geq c\\
    0 & |x|< c\\
  \end{array} \right.
\label{eq:eq10}
\end{equation}
where $c$ is a fixed threshold. 

In summary, Fu and Wei considered that central pixels are more important than their neighbors and thus proposed the centralized LBP ($LBP^{cen}_{P,R}$) as follows:
\begin{equation}
    LBP^{cen}_{P, R}\left(g_{c}\right) = \sum_{p=0}^{\frac{p}{2}-1}{s\left(g_{p}-g_{p+\frac{p}{2}}\right)2^{p}} + s\left(g_{c}-g_{tot}\right)2^{\frac{p}{2}}
\label{eq:eq11}
\end{equation}
and $g_{tot}$ is defined as:
\begin{equation}
      g_{tot} = \frac{1}{p+1}\left(g_{c}+\sum_{p=0}^{p-1}{g_{p}}\right)
\label{eq:eq12}
\end{equation}
where $g_{c}$ and $\left(x_{c}, y_{c}\right)$ represent the intensity and coordinates of the central pixel respectively. Due to the fact that the algorithm considers correlation between opposite pixel points, this algorithm is not rotational invariant.

\cite{TAN2010a} proposed an extension to the operator, Eq.~(\ref{eq:eq1}), called extended LBP ($LBP^{ext}_{P,R}$) by using the value of central pixels plus a tolerance interval $t$ as local threshold, $t$ is a user-specific value, usually set at ``1''. Each pixel value within the interval zone $g_{c} \pm t$ is quantized as zero. Pixel values above the tolerance interval are labeled with ``$1$'' and those below this zone are labeled with ``$-1$'' as follows:
\begin{equation}
   s\left(x\right) = \left\{
  \begin{array}{lc}
    1 & \mbox{if $x > t$}\\
    0 & \mbox{if $\left|x\right| \leq t$}\\
    -1 & \mbox{if $x < -t$}
  \end{array} \right.
\label{eq:eq13}
\end{equation}
here $x$ is the difference between the $P$ neighbors and their central pixels. Each ternary pattern is split into upper and lower pattern and each part is encoded as a separate LBP pattern. Finally, their LBP histograms are concatenated.

\cite{GUO2010a} suggested using both the sign and magnitude of a $d_{p}$ vector to form the so-called completed LBP ($LBP^{com}_{P,R}$). In Eq.~(\ref{eq:eq1}) only the sign component is considered whereas in the Guo et al. proposal, $d_{p} = \left\{g_{p}-g_{c} | p = 0, 1, \ldots, P-1\right\}$ is split into two components as follows:
\begin{equation}
   d_{p} = s_{p}\ast m_{p} = \left\{
  \begin{array}{ll}
    s_{p} &= sign\left(d_{p}\right)\\
    m_{p} &= |d_{p}|
  \end{array} \right.
\label{eq:eq14}
\end{equation}
where $s_{p}$ and $m_{p}$ are the sign and magnitude of $d_{p}$ respectively. In addition, they presented an analysis of the sign component and concluded that $s_{p}$ preserves more information of $d_{p}$ than $m_{p}$. They defined three operators that are combined to build the LBP histogram: $LBP^{com}_{S_{P,R}}$, which considers the sign component of $d_{p}$, $LBP^{com}_{C_{P,R}}$ which considers the magnitude component of $d_{p}$, and $LBP^{com}_{C_{P,R}}$ which considers the magnitude of central pixels.

Finally,~\cite{LIAO2009} proposed the dominant LBP ($LBP^{dom}_{P,R}$) which is a modification of Eq.~(\ref{eq:eq5}) based on the fact that $LBP^{uni}_{P,R}$ in practice is not well suited to encode some complicated pattern textures such as curvature edges and crossing boundaries of corners. A possible explanation is due to the fact that the extracted uniform patterns do not have a dominant proportion of them to better represent the object (or image). Liao et al. have shown that given a set of training images, the required number of patterns to better representing textures corresponds to at least $80\%$ of the pattern occurrences. The first step of their procedure is to compute the LBP histogram and sort it in descending order. The second step is to extract a vector for obtain $80\%$ of pattern occurrences. This procedure guarantees a suitable framework for representing textures.

%-------------------------------------------------------------------------
\subsection{LBP histogram evaluation~\label{sec:distance}}
Since LBP operators act as fingerprint of texture, it is possible to use LBP histogram distances as a similarity measure among all different textures. Two different methodologies can be used for histogram distance evaluation: vector and probabilistic approaches. In the vector approach, a histogram is treated as a fixed-dimensional vector. Hence standard vector norms such as city block or Euclidean between univariate histograms can be used. The probabilistic approach is based on the fact that a histogram provides the basis for an empirical estimation of the probabilistic density function (pdf). Computing the distance between two histograms is equivalent to measure the overlapping part between two pdf's as the distance. 

Although the Kullback-Leibler divergence (KL) --a generalization of Shannon's entropy-- is not a true metric rather it is a relative entropy, it can be used as a suitable metric for measuring distances between histograms as follows:
\begin{equation}
D_{KL}\left(A, B\right) = \sum_{i = 0}^{b - 1}{P_{i}\left(B\right)\log\frac{P_{i}\left(B\right)}{P_{i}\left(A\right)}}
\label{eq:eq15}
\end{equation}
where $A$ and $B$ are two histograms with $b$ bins length each, and $P_{i}$ denotes the probability of the bin $i$.

\cite{CHA2002} proposed a novel histogram distance measure called ordinal distance (OD) based on the idea that a histogram $h\left(A\right)$ can be transformed into a histogram $h\left(B\right)$ by moving elements from left to right, being the total movements the distance between them.

\begin{equation}
D_{ord}\left\{h\left(A\right), h\left(B\right)\right\} = \sum_{i = 0}^{b - i}{\left|\sum_{j = 0}^{i}{\left(h_{j}\left(A\right), h_{j}\left(B\right)\right)} \right|}
\label{eq:eq16}
\end{equation}

In the next Section~\ref{sec:experiments}, we present several experiments of texture classification using both KL and OD metrics in order to compare the performance of seven LBP approaches under several settings.

%-------------------------------------------------------------------------
\section{Experiments and results~\label{sec:experiments}}
\begin{figure}[htbp]
    \centering
    \subfigure{\includegraphics[width=0.19\textwidth]{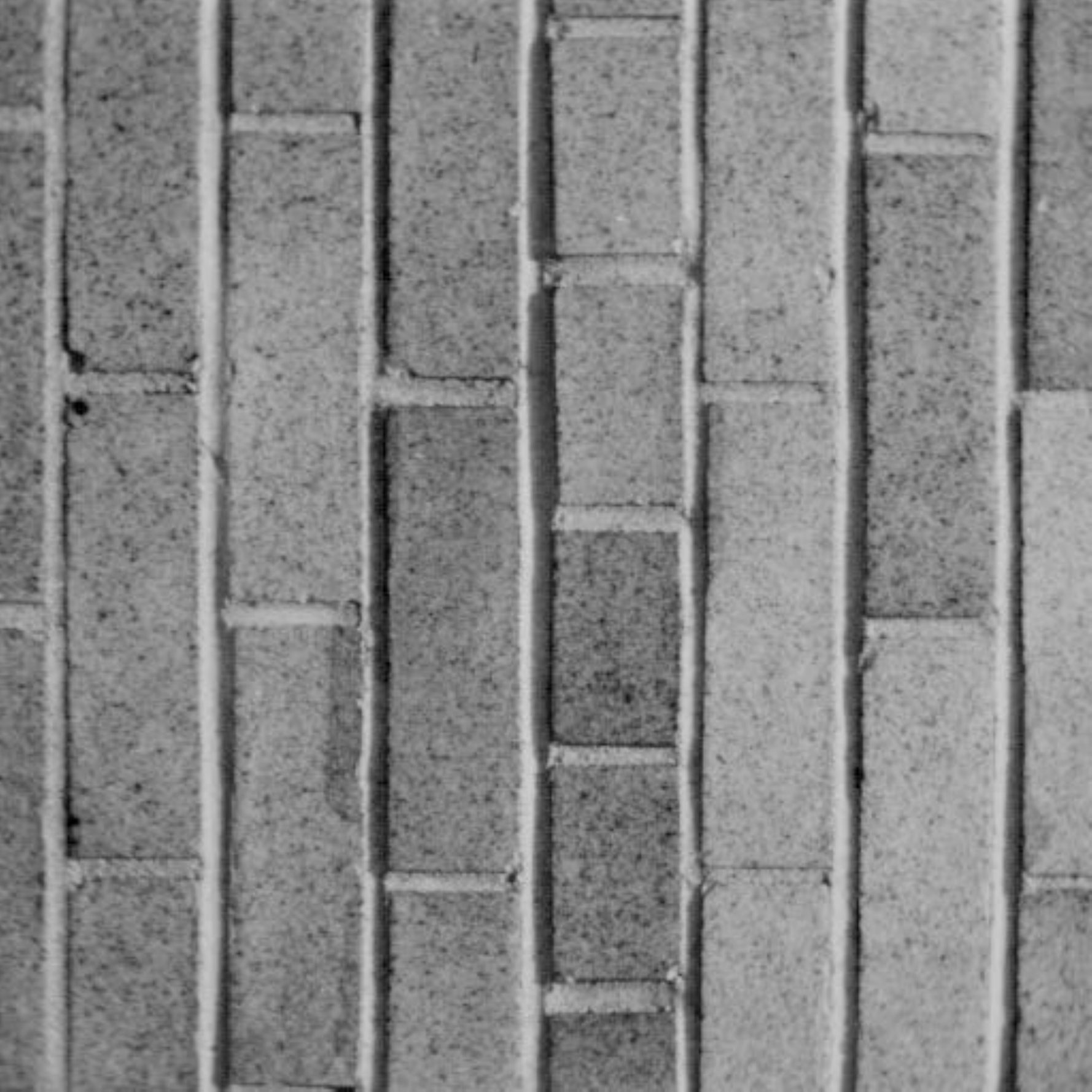}}
    \subfigure{\includegraphics[width=0.19\textwidth]{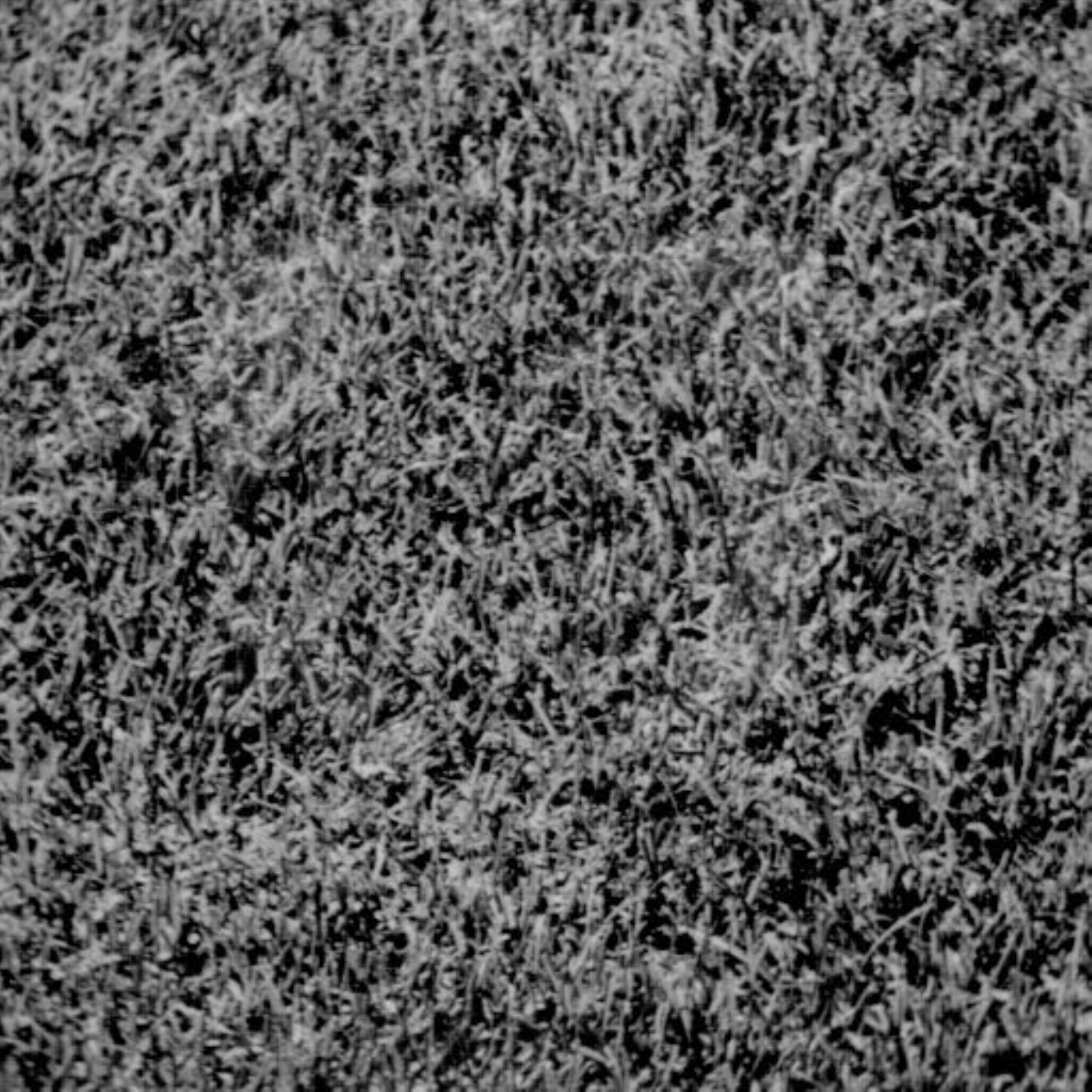}}
    \subfigure{\includegraphics[width=0.19\textwidth]{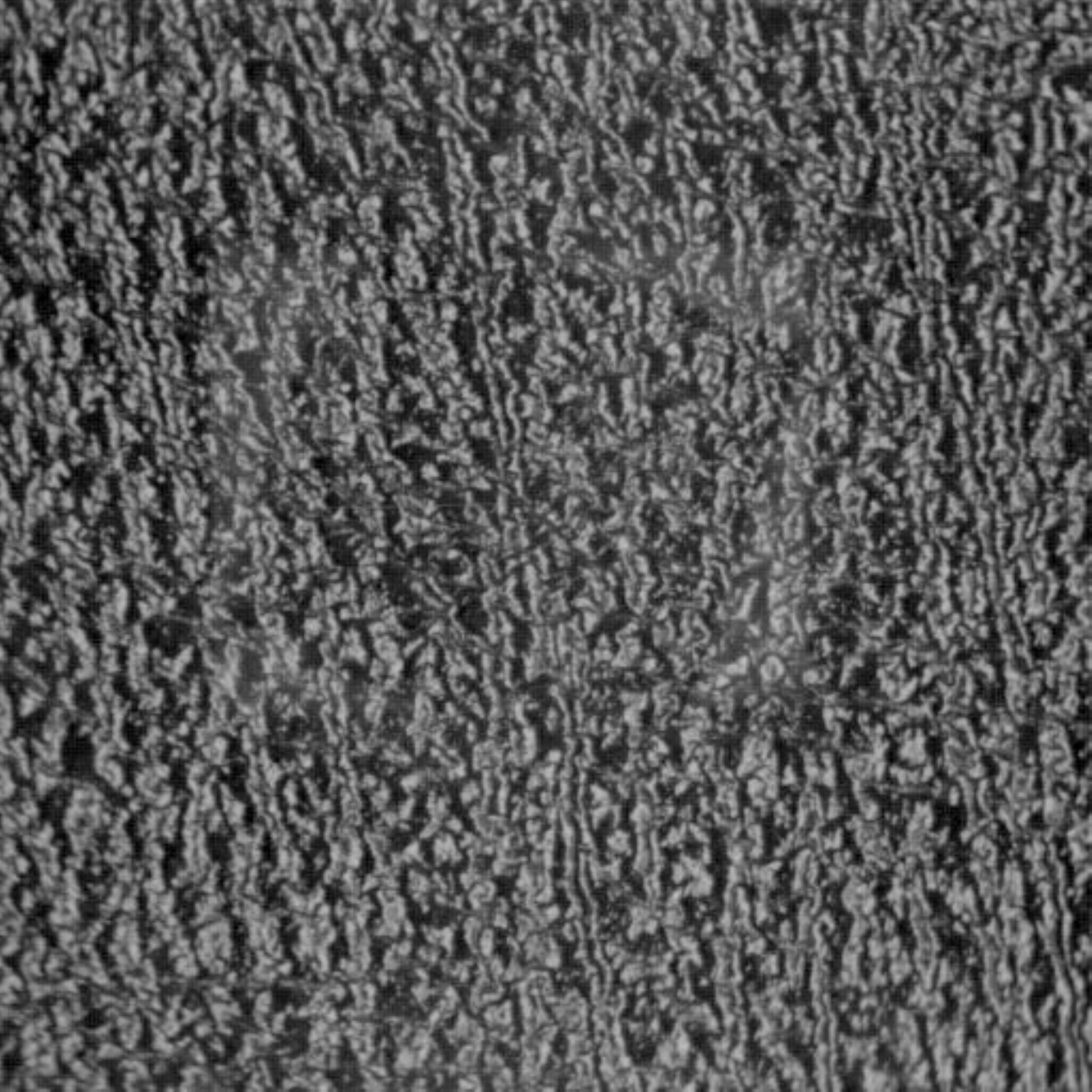}}
    \subfigure{\includegraphics[width=0.19\textwidth]{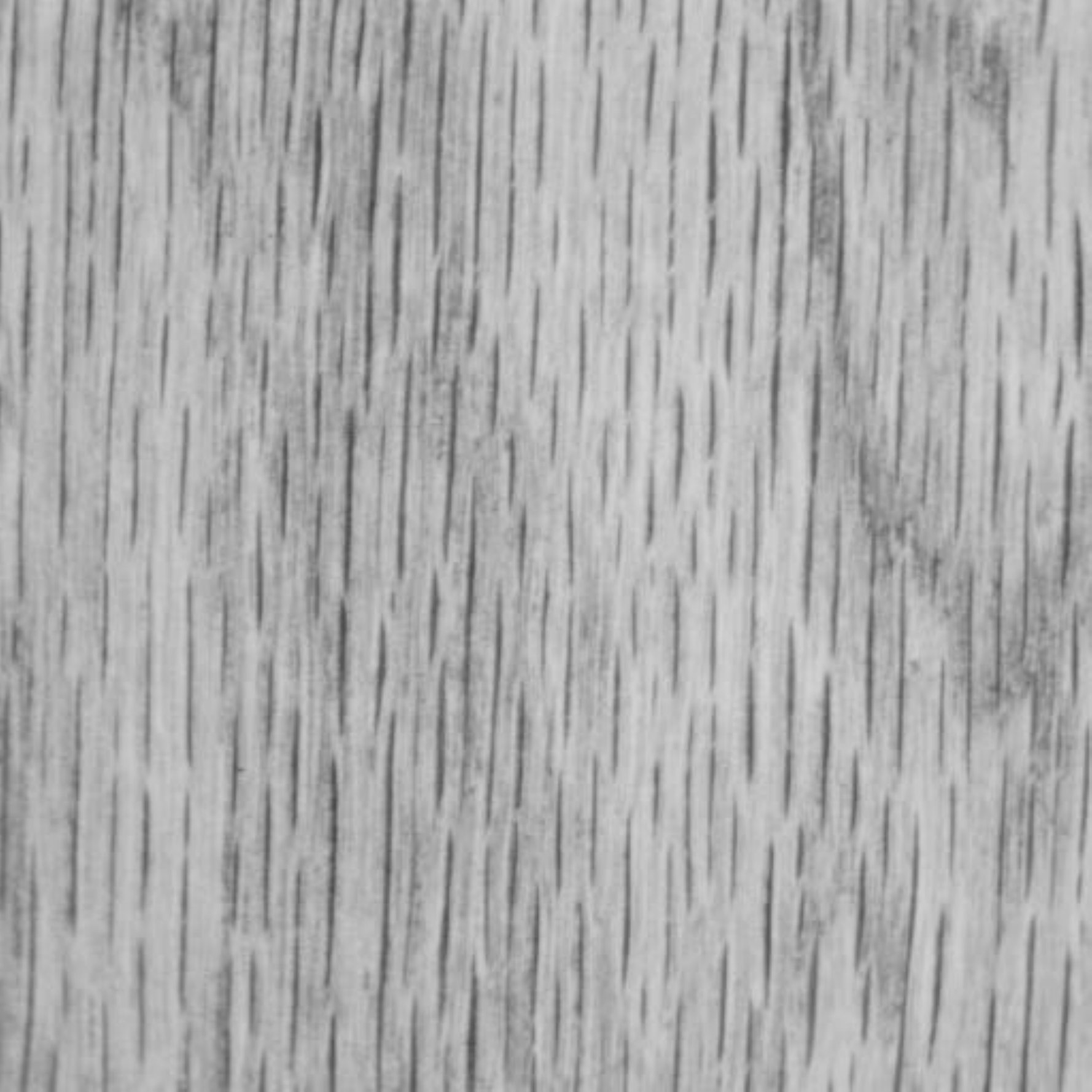}}
    \subfigure{\includegraphics[width=0.19\textwidth]{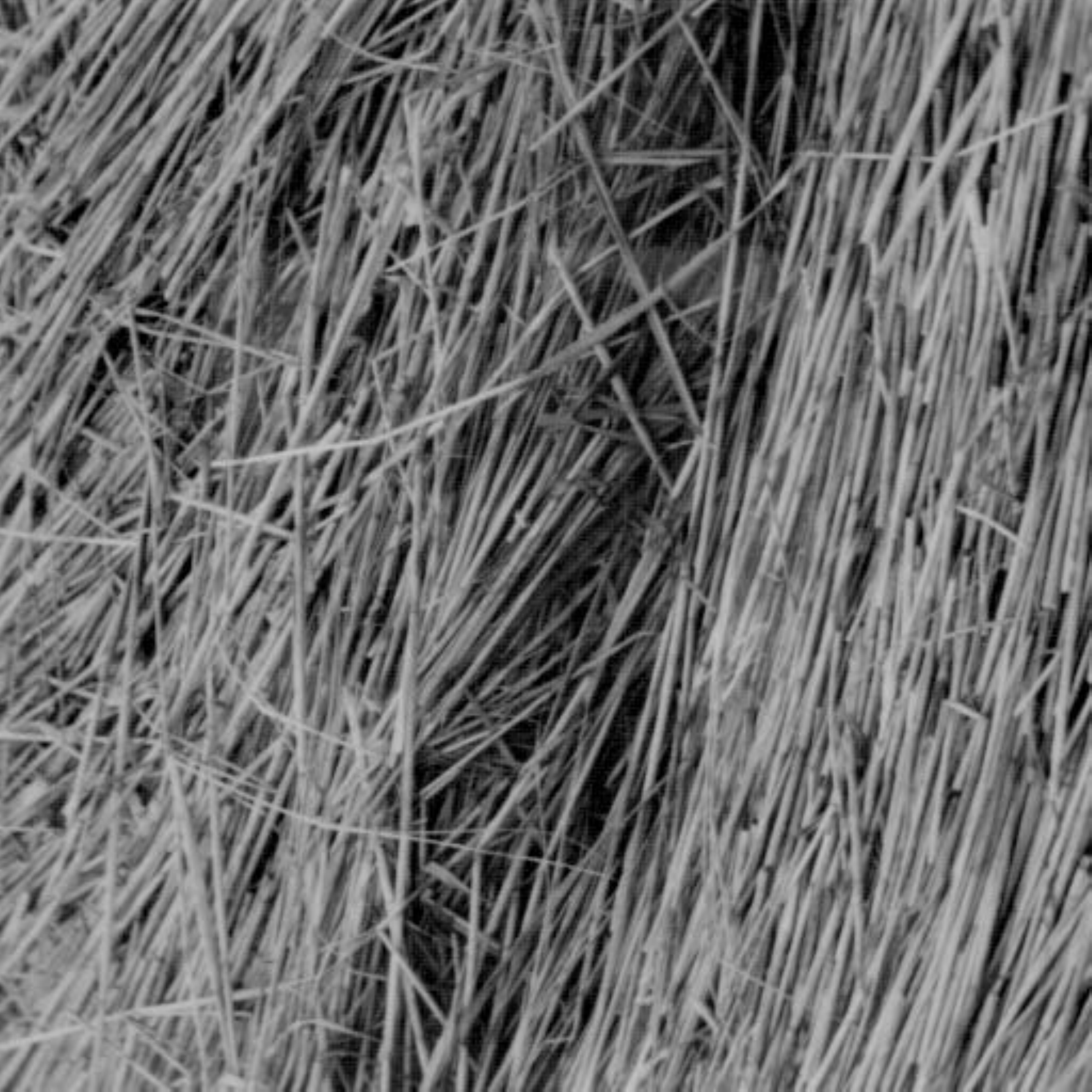}}\\
    \subfigure{\includegraphics[width=0.19\textwidth]{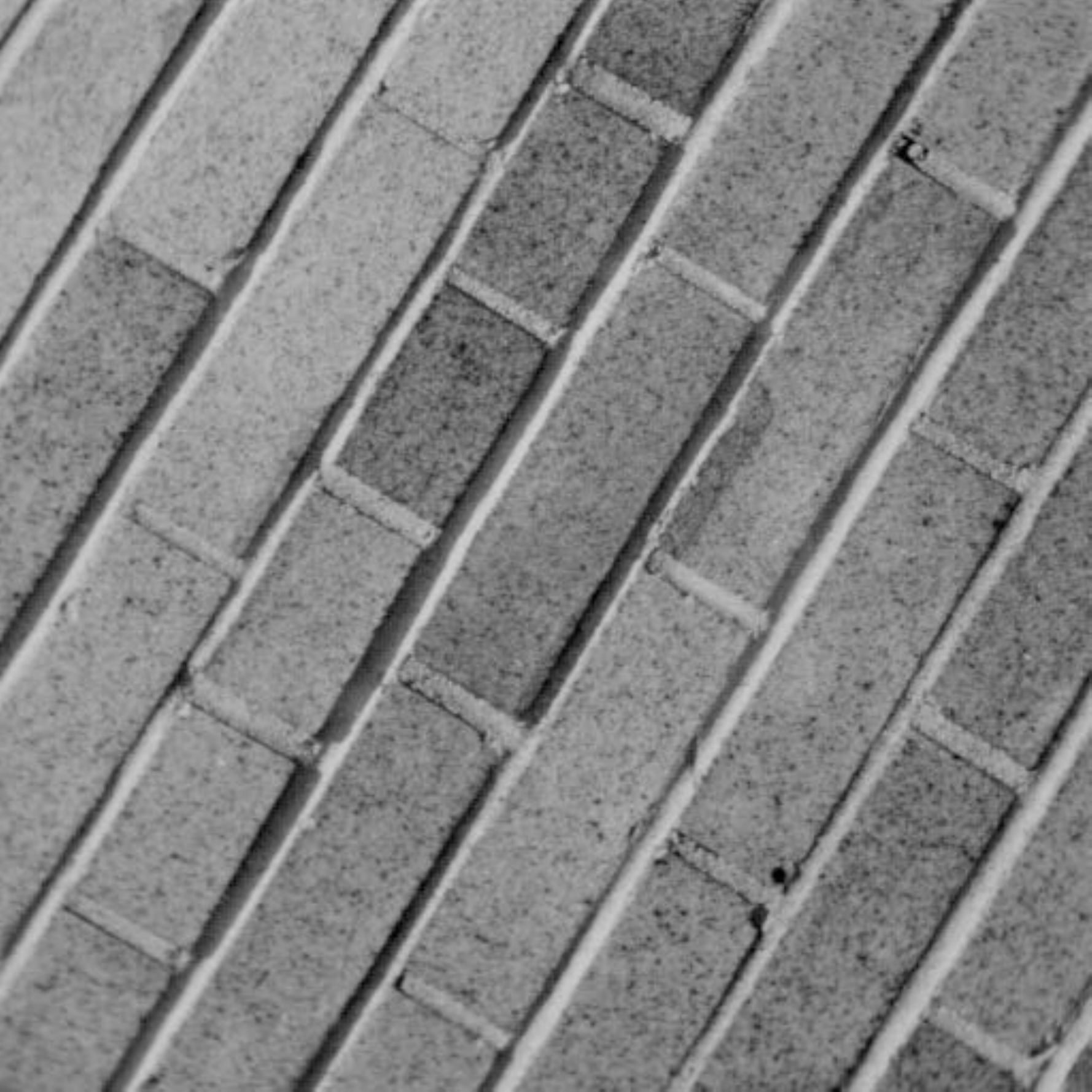}}
    \subfigure{\includegraphics[width=0.19\textwidth]{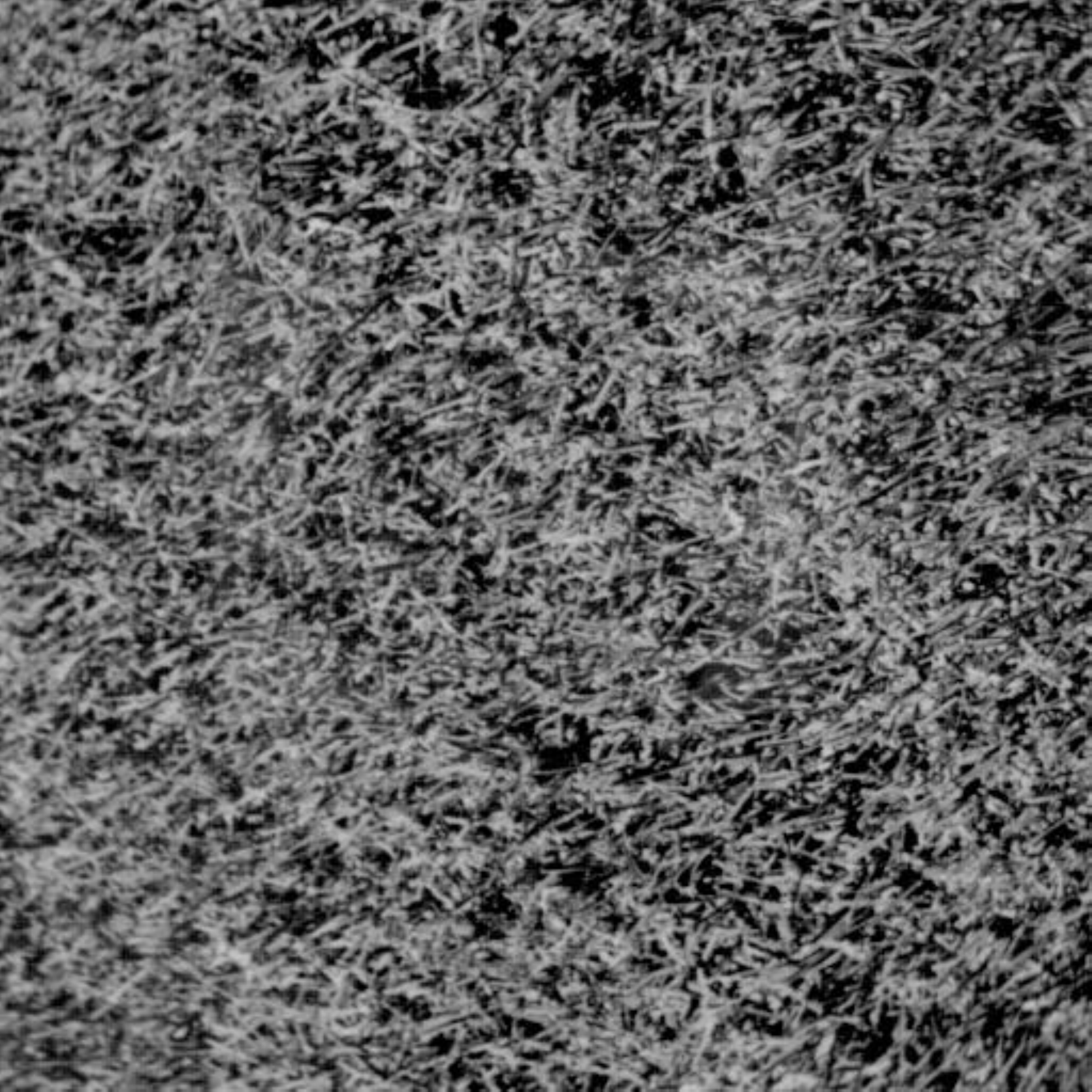}}
    \setcounter{subfigure}{0}\subfigure[]{\label{fig:fig2_a}\includegraphics[width=0.19\textwidth]{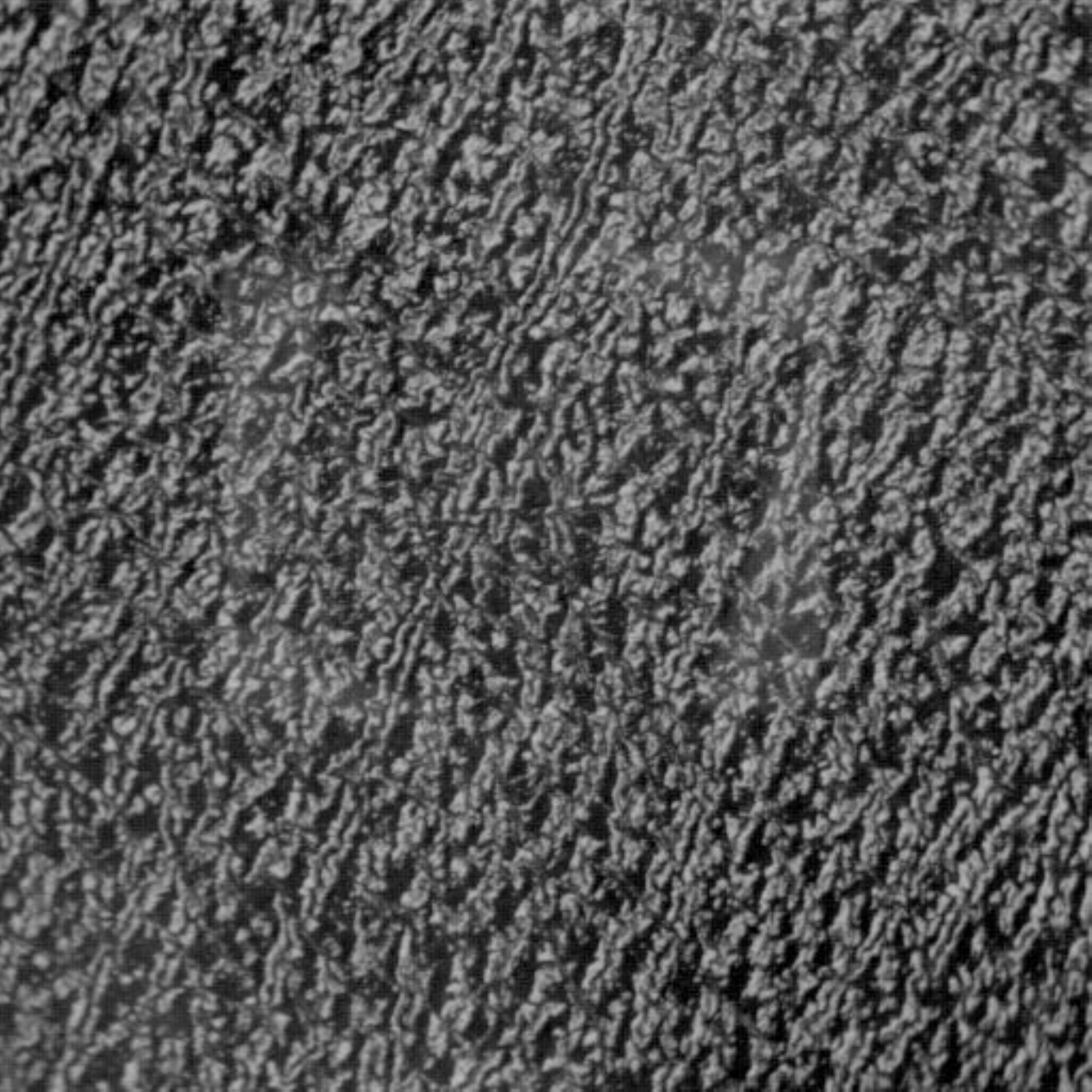}}
    \subfigure{\includegraphics[width=0.19\textwidth]{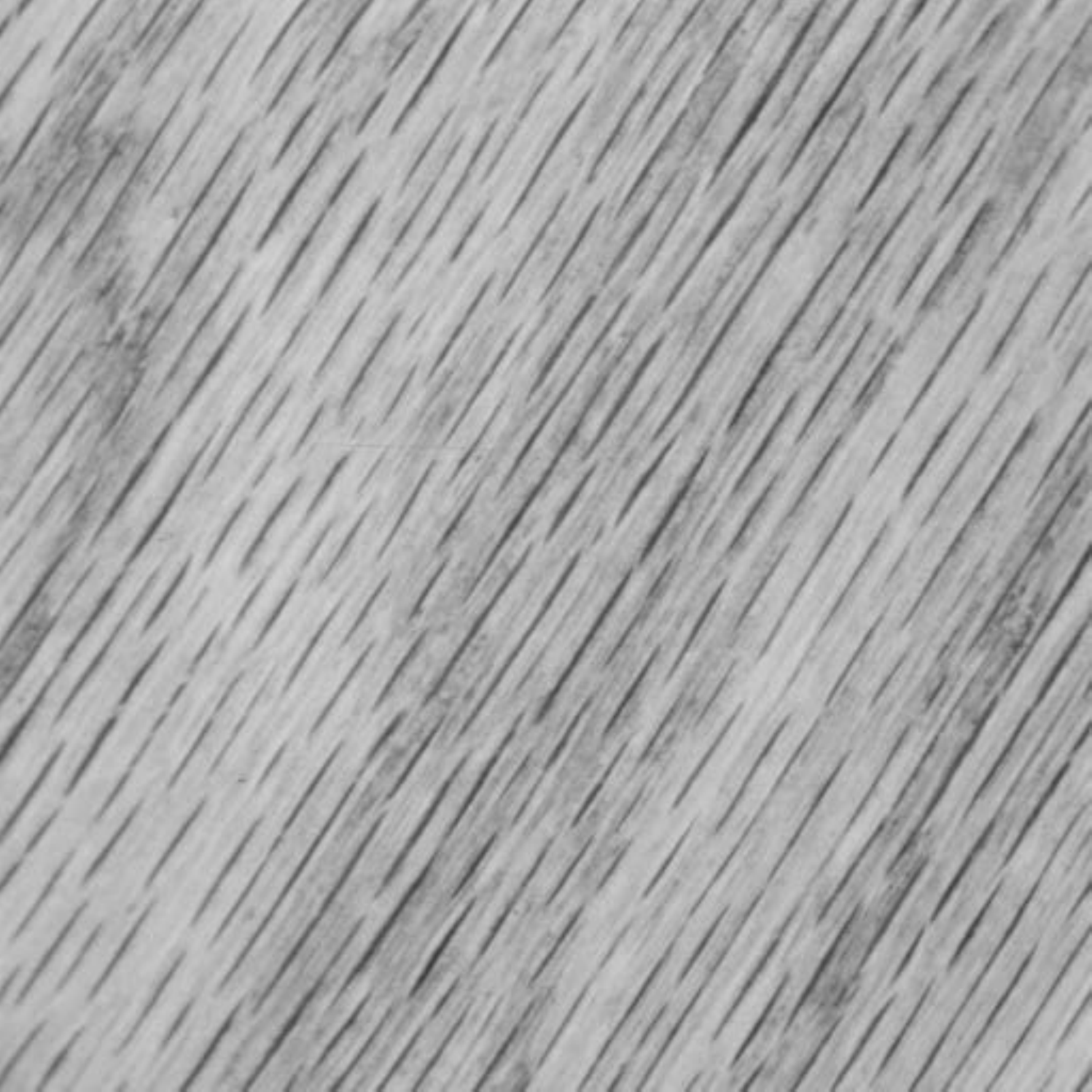}}
    \subfigure{\includegraphics[width=0.19\textwidth]{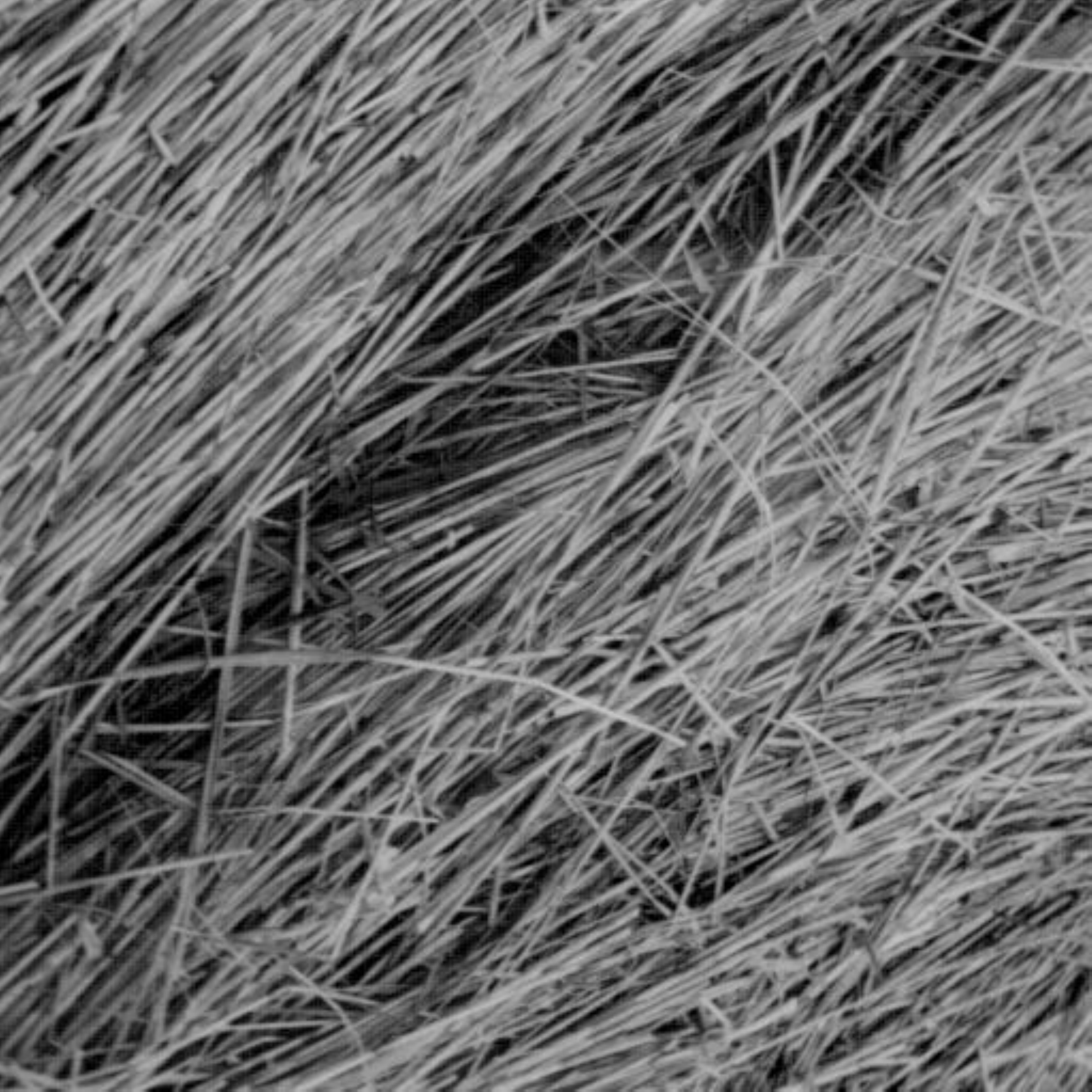}}\\
    \subfigure{\includegraphics[width=0.19\textwidth]{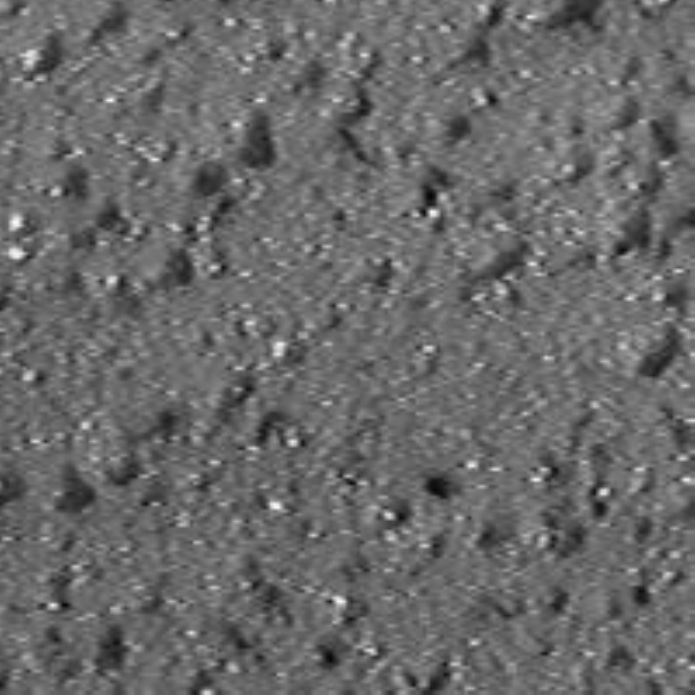}}
    \subfigure{\includegraphics[width=0.19\textwidth]{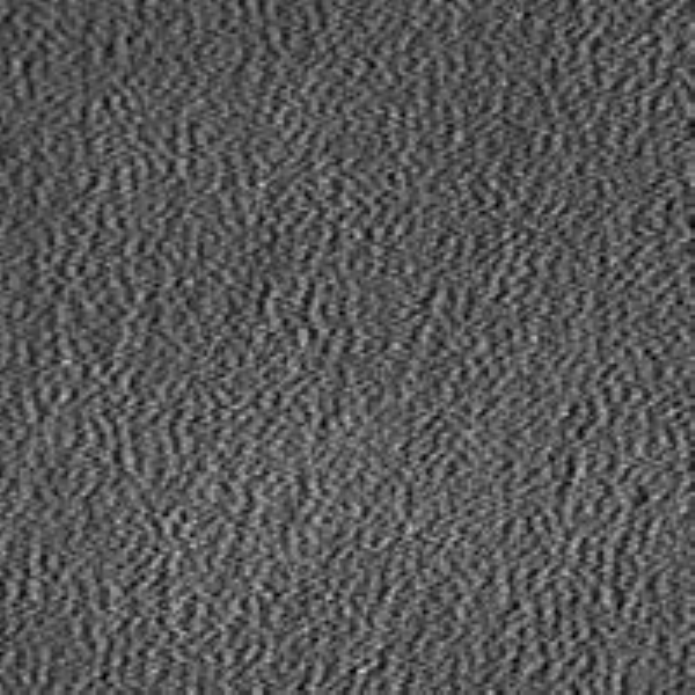}}
    \subfigure{\includegraphics[width=0.19\textwidth]{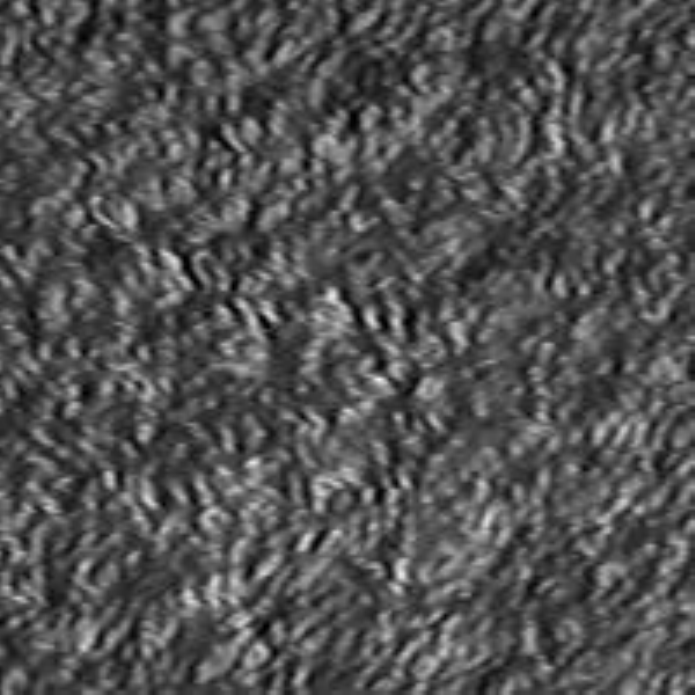}}
    \subfigure{\includegraphics[width=0.19\textwidth]{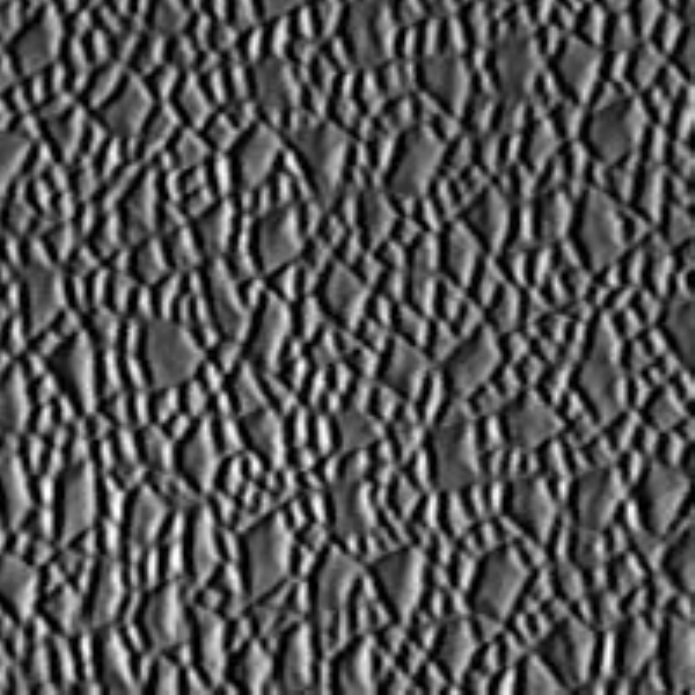}}
    \subfigure{\includegraphics[width=0.19\textwidth]{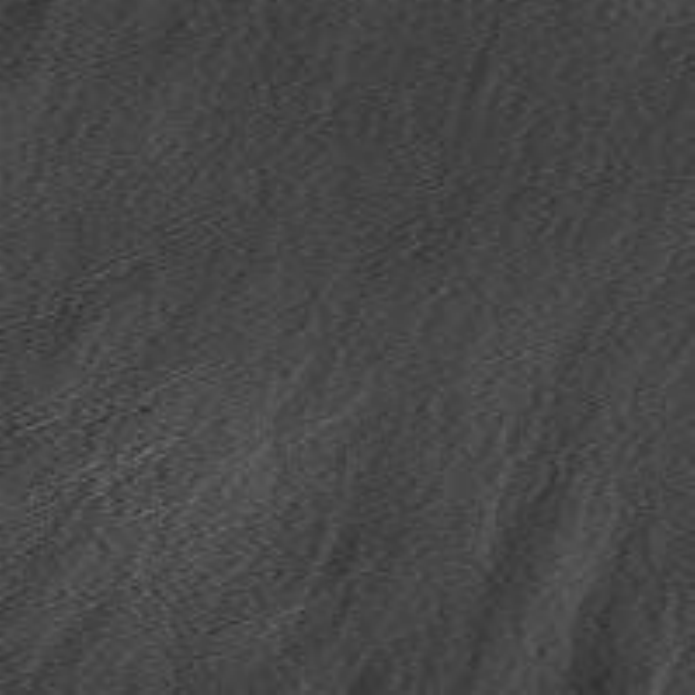}}\\
    \subfigure{\includegraphics[width=0.19\textwidth]{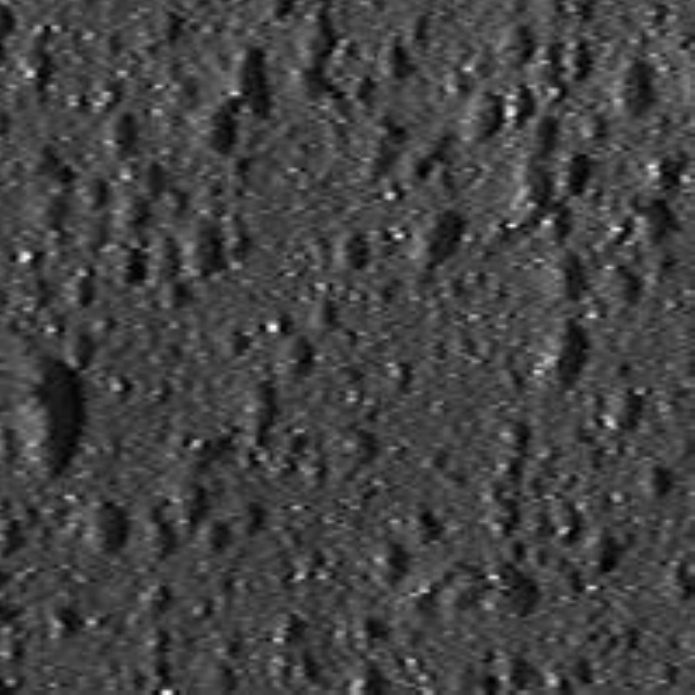}}
    \subfigure{\includegraphics[width=0.19\textwidth]{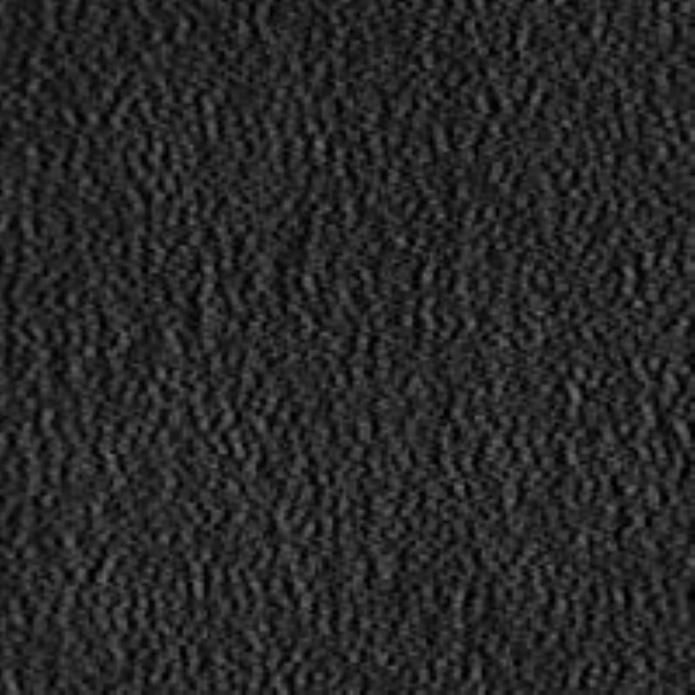}}
    \setcounter{subfigure}{1}\subfigure[]{\label{fig:fig2_b}\includegraphics[width=0.19\textwidth]{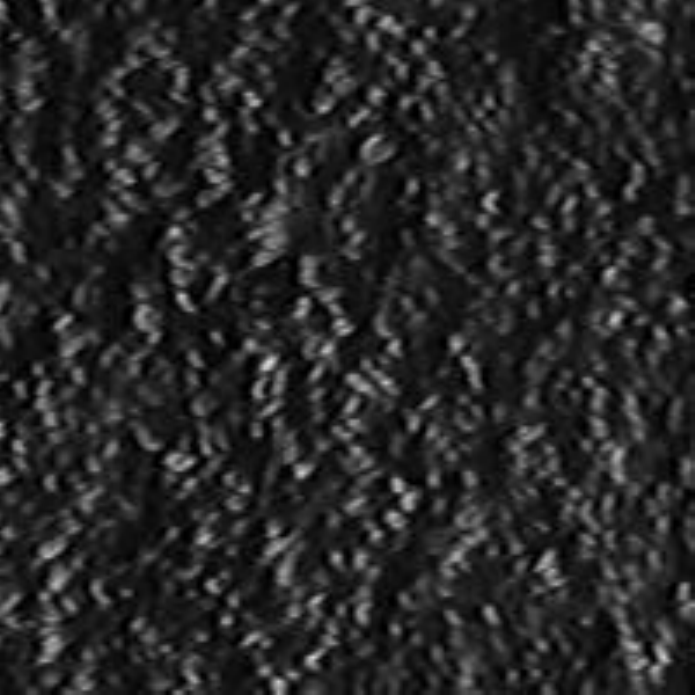}}
    \subfigure{\includegraphics[width=0.19\textwidth]{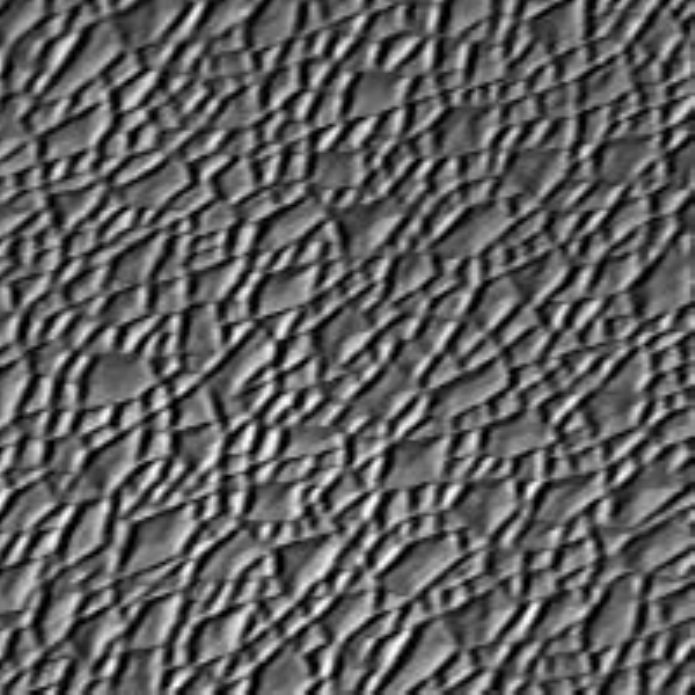}}
    \subfigure{\includegraphics[width=0.19\textwidth]{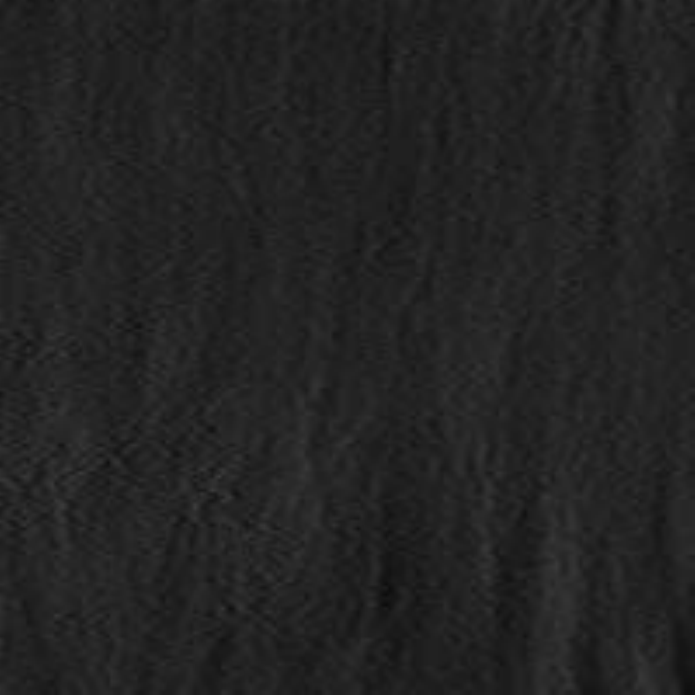}}\\
    \caption{\ref{fig:fig2_a} The first row shows five reference images: brick (D94), grass (D9), leather (D24), wood (D68), and straw (D15), the second row shows the rotated versions of the reference images, ($150^{\circ}$). \ref{fig:fig2_b}. The third and the fourth rows present a subset of images from the CUReT database.}
    \label{fig:fig2} 
\end{figure}

We split experimental assessments into three categories: rotational, noisy, and illumination changes. The first two evaluations were performed using the USC-SIPI image database available at \cite{USC-SIPI}. This database is a rotated version of Brodatz database and consists of thirteen images each digitized at seven different rotation angles: $0$, $30$, $60$, $90$, $120$, $150$, and $200$ degrees. The images are all $512\times 512$ pixels with $8$ bits/pixel, see Fig.~\ref{fig:fig2_a}. 

For illumination tests we used a large enough subset of the CUReT database, available at \cite{CUReT}, to validate our experiments. In the last case, we employed ten reference images with ten illumination variations each for a total of $100$ images; the images are all $200 \times 200$ pixels and were converted into gray scale with $8$ bits/pixel, see Fig.~\ref{fig:fig2_b}. The classification procedure setup consisted of comparing LBP histogram distances of each reference image against the test images.

In order to evaluate the performance of the previously described seven rotational invariant LBP approaches, we used the next thirteen textures as reference images: bark (D12), brick (D94), bubbles (D112), grass (D9), leather (D24), pigskin (D92), raffia (D84), sand (D29), straw (D15), water(D38), weave (D16), wood (D68), and wool (D19) -the number between parenthesis is the identification number in the Brodatz texture book, \cite{BRODATZ1966}. We measured and compared distances among all histograms ($71$ rotated textures and $13$ reference textures) using the two metrics previously presented in Eq.~(\ref{eq:eq15}) and Eq.~(\ref{eq:eq16}) and we assigned a rotated image to a certain texture according to the closest distance between LBP histograms. See Table~\ref{tab:tab1} for comparative study of the accuracy performance. In addition we used confusion matrices to obtained the accuracy rate (AR) of the seven LBP approaches using the next equation:
\begin{equation}
    AR = \left(\frac{\sum_i^k{a_{i,i}}}{\sum_{i,j}^k{a_{i,j}}}\right)\times 100\%
    \label{eq:eq17}
\end{equation}
where $\left(i,j\right)$ are matrix indexes and $k$ is the number of texture references.

\begin{table}[htbp]
\setlength{\tabcolsep}{3pt}
\renewcommand{\arraystretch}{1.3}
\caption{Comparison of seven LBP approaches. $LBP^{min}$ and $LBP^{min}_{P,R}$ differ that the first one does not use interpolated neighbors but the second one does.}
\label{tab:tab1}
\centering
\scriptsize
\begin{tabular}{cccccr@{.}}
\hline
 \multirow{2}{*}{Scheme} & \multicolumn{2}{c}{OD metric} & \multicolumn{2}{c}{KLD metric} & \multirow{2}{*}{Reference} \\
\cline{2-5}
 & $\#$ textures & Accuracy rate $\left(\%\right)$ & $\#$ textures  & Accuracy rate  $\left(\%\right)$ & \\
\hline
 $LBP$ & 35 & 38.46 & 39 & 42.86 & \cite{OJALA1994} \\
 $LBP^{min}$ & 79 & 86.81 & 84 & 92.31 & \cite{OJALA1994}\\
 $LBP^{min}_{P,R}$ & 77 & 84.62 & 72 & 74.00 & \cite{PIETIKAINEN2000} \\
 $LBP^{uni}_{P,R}$ & 80 & 87.91 & 82 & 90.11 & \cite{OJALA2002} \\
 $LBP^{num}_{P,R}$ & 83 & 91.21 & 80 & 87.91 & \cite{MA2011} \\
 $LBP^{ni}_{P,R}$ & 76 & 83.52 & 74 & 81.32 & \cite{LIU2011} \\
 $LBP^{med}_{P,R}$ & 72 & 79.12 & 64 & 70.33 & \cite{ZABIH1994}\\
\hline
\end{tabular}
\end{table}

Original Ojala's proposal achieved the lowest AR because is not invariant to rotation with $35$ out of $91$ textures correctly classified in the worst scenario. Although this paper is focused to analyze invariant to rotation approaches, we performed a comparison between $LBP$ and $LBP^{cen}_{P,R}$, (see Eq.~\ref{eq:eq11}). $LBP^{cen}_{P,R}$ achieved an AR of $29.61\%$ using the OD metric and $36.26\%$ of textures correctly classified under the KL metric. $LBP^{cen}_{P,R}$ performance was even lower than original Ojala's proposal. One of the possible reasons is that $LBP^{cen}_{P,R}$ uses a fixed threshold $c$ in Eq.~(\ref{eq:eq10}) which influences the accuracy rate.

On the other hand, $LBP^{num}_{P,R}$ achieved $91.91\%$ of textures correctly classified, $3.3\%$ more than $LBP^{uni}_{P,R}$ with $87.91\%$. This can be interpreted as $LBP^{num}_{P,R}$ add extra information of non-uniform patterns into the LBP histogram whereas $LBP^{uni}_{P,R}$ labels all non-uniform patterns into a unique label which discards the large amount of texture information represented by these patterns. Another possible explanation is that stochastic components are disregarded when all non-uniform patterns are grouped because they represent abrupt variations and changes in the textures.

Table~\ref{tab:tab2} and Table~\ref{tab:tab3} present the best AR in the form of confusion matrices for the OD and KLD metrics respectively.

\begin{table}[htbp]
\setlength{\tabcolsep}{2pt}
\renewcommand{\arraystretch}{1.3}
\caption{Confusion matrix for the classification experiment of $LBP^{num}_{P,R}$ using OD metric. Major mistakes occurred between grass (D9) and leather (D24) textures and between wood (D68) and straw (D15) textures.}
\label{tab:tab2}
\centering
\scriptsize
\begin{tabular}{ccccccccccccccc}
\hline
 & & \multicolumn{13}{c}{Predicted} \\
\cline{3-15}
 & & bark & brick & bubbles & grass & leather & pigskin & raffia & sand & straw & water & weave & wood & wool \\
\cline{3-15}
\multirow{13}{*}{\rotatebox{90}{Actual}} & bark & 7 & 0 & 0 & 0 & 0 & 0 & 0 & 0 & 0 & 0 & 0 & 0 & 0 \\
 \cline{2-15}
 & brick & 0 & 7 & 0 & 0 & 0 & 0 & 0 & 0 & 0 & 0 & 0 & 0 & 0 \\
 \cline{2-15}
 & bubbles & 0 & 0 & 7 & 0 & 0 & 0 & 0 & 0 & 0 & 0 & 0 & 0 & 0 \\
 \cline{2-15}
 & grass & 0 & 0 & 0 & 4 & 3 & 0 & 0 & 0 & 0 & 0 & 0 & 0 & 0 \\
 \cline{2-15}
 & leather & 0 & 0 & 0 & 0 & 7 & 0 & 0 & 0 & 0 & 0 & 0 & 0 & 0 \\
 \cline{2-15}
 & pigskin & 0 & 0 & 0 & 0 & 0 & 7 & 0 & 0 & 0 & 0 & 0 & 0 & 0 \\
 \cline{2-15}
 & raffia & 0 & 0 & 0 & 0 & 0 & 0 & 7 & 0 & 0 & 0 & 0 & 0 & 0 \\
 \cline{2-15}
 & sand & 2 & 0 & 0 & 0 & 0 & 0 & 0 & 7 & 0 & 0 & 0 & 0 & 0 \\
 \cline{2-15}
 & straw & 0 & 0 & 0 & 0 & 0 & 0 & 0 & 0 & 7 & 0 & 0 & 0 & 0 \\
 \cline{2-15}
 & water & 0 & 0 & 0 & 0 & 0 & 0 & 0 & 0 & 0 & 7 & 0 & 0 & 0 \\
 \cline{2-15}
 & weave & 0 & 0 & 0 & 0 & 0 & 0 & 0 & 0 & 0 & 0 & 7 & 0 & 0 \\
 \cline{2-15}
 & wood & 0 & 0 & 0 & 0 & 0 & 0 & 0 & 0 & 5 & 0 & 0 & 2 & 0 \\
 \cline{2-15}
 & wool & 0 & 0 & 0 & 0 & 0 & 0 & 0 & 0 & 0 & 0 & 0 & 0 & 7 \\
\hline
\end{tabular}
\normalsize
\end{table}

From Table~\ref{tab:tab3} one can observe that $LBP^{min}$ provides the higher rate using the KL metric. Since KL metric is a measure of relative entropy, there is a strong suspicious that when neighbors are calculated by bilinear interpolation, extra information is added which increases the distance between LBP histograms affecting the classification.

\begin{table}[htbp]
\setlength{\tabcolsep}{2pt}
\renewcommand{\arraystretch}{1.3}
\caption{Confusion matrix for the classification experiment of $LBP^{min}$ using KLD metric. Most mistakes occurred when rotated wool textures were classified as straw texture. On the contrary, all the rotated straw textures were correctly classified.}
\label{tab:tab3}
\centering
\scriptsize
\begin{tabular}{ccccccccccccccc}
\hline
 & & \multicolumn{13}{c}{Predicted} \\
\cline{3-15}
 & & bark & brick & bubbles & grass & leather & pigskin & raffia & sand & straw & water & weave & wool & wood \\
\cline{3-15}
\multirow{13}{*}{\rotatebox{90}{Actual}} & bark & 7 & 0 & 0 & 0 & 0 & 0 & 0 & 0 & 0 & 0 & 0 & 0 & 0 \\
 \cline{2-15}
 & brick & 0 & 7 & 0 & 0 & 0 & 0 & 0 & 0 & 0 & 0 & 0 & 0 & 0 \\
 \cline{2-15}
 & bubbles & 0 & 0 & 7 & 0 & 0 & 0 & 0 & 0 & 0 & 0 & 0 & 0 & 0 \\
 \cline{2-15}
 & grass & 0 & 0 & 0 & 6 & 1 & 0 & 0 & 0 & 0 & 0 & 0 & 0 & 0 \\
 \cline{2-15}
 & leather & 0 & 0 & 0 & 0 & 7 & 0 & 0 & 0 & 0 & 0 & 0 & 0 & 0 \\
 \cline{2-15}
 & pigskin & 0 & 0 & 0 & 0 & 0 & 7 & 0 & 0 & 0 & 0 & 0 & 0 & 0 \\
 \cline{2-15}
 & raffia & 0 & 0 & 0 & 0 & 0 & 0 & 7 & 0 & 0 & 0 & 0 & 0 & 0 \\
 \cline{2-15}
 & sand & 0 & 0 & 0 & 0 & 0 & 0 & 0 & 7 & 0 & 0 & 0 & 0 & 0 \\
 \cline{2-15}
 & straw & 0 & 0 & 0 & 0 & 0 & 0 & 0 & 0 & 7 & 0 & 0 & 0 & 0 \\
 \cline{2-15}
 & water & 0 & 0 & 0 & 0 & 0 & 1 & 0 & 0 & 0 & 6 & 0 & 0 & 0 \\
 \cline{2-15}
 & weave & 0 & 0 & 0 & 0 & 0 & 0 & 0 & 0 & 0 & 0 & 7 & 0 & 0 \\
 \cline{2-15}
 & wool & 0 & 0 & 0 & 0 & 0 & 0 & 0 & 0 & 5 & 0 & 0 & 2 & 0 \\
 \cline{2-15}
 & wood & 0 & 0 & 0 & 0 & 0 & 0 & 0 & 0 & 0 & 0 & 0 & 0 & 7 \\
\hline
\end{tabular}
\normalsize
\end{table}

\begin{figure}[htbp]
    \centering
    \subfigure[]{\label{fig:fig_3a}\includegraphics[width=0.85\textwidth]{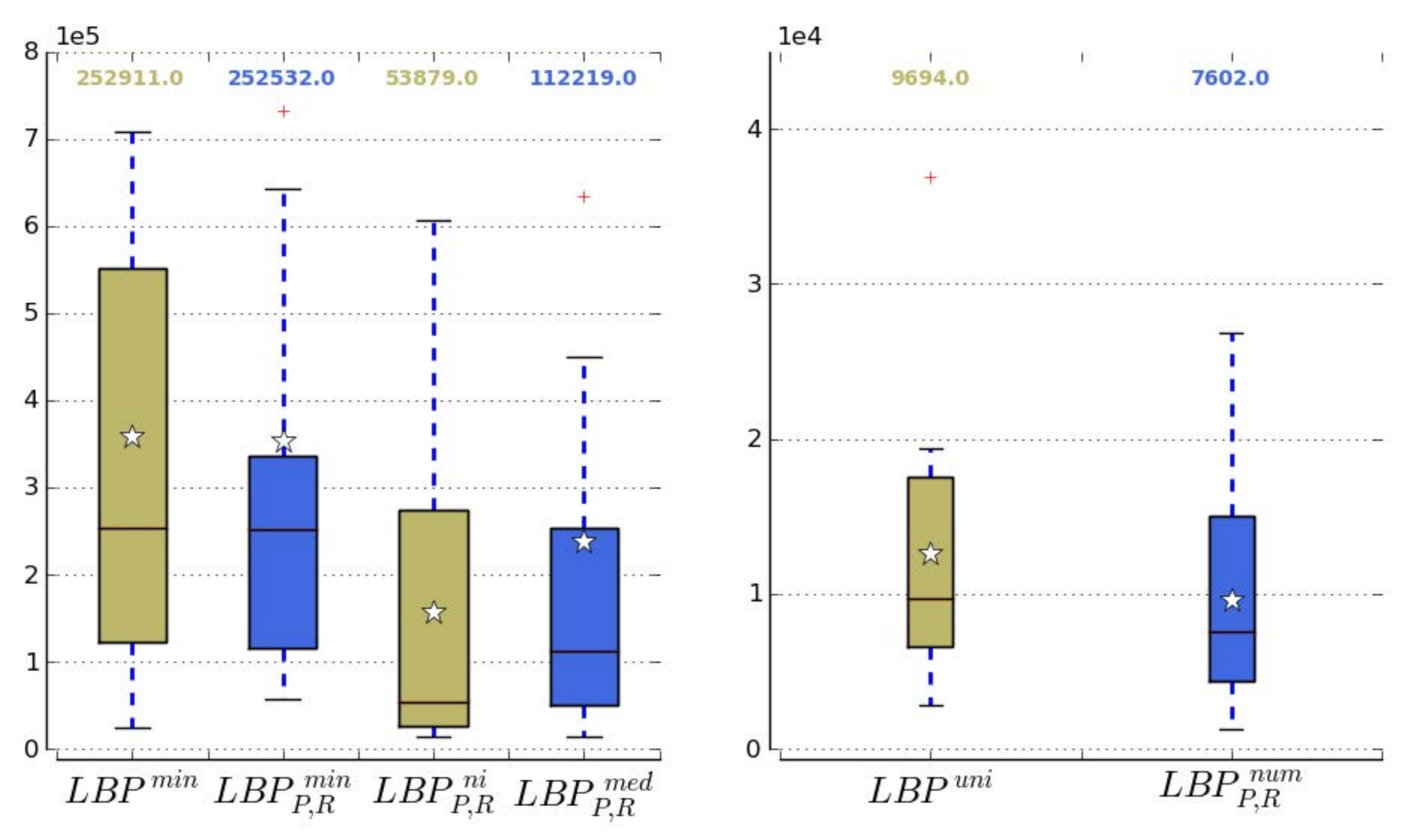}}\\
    \subfigure[]{\label{fig:fig_3b}\includegraphics[width=0.85\textwidth]{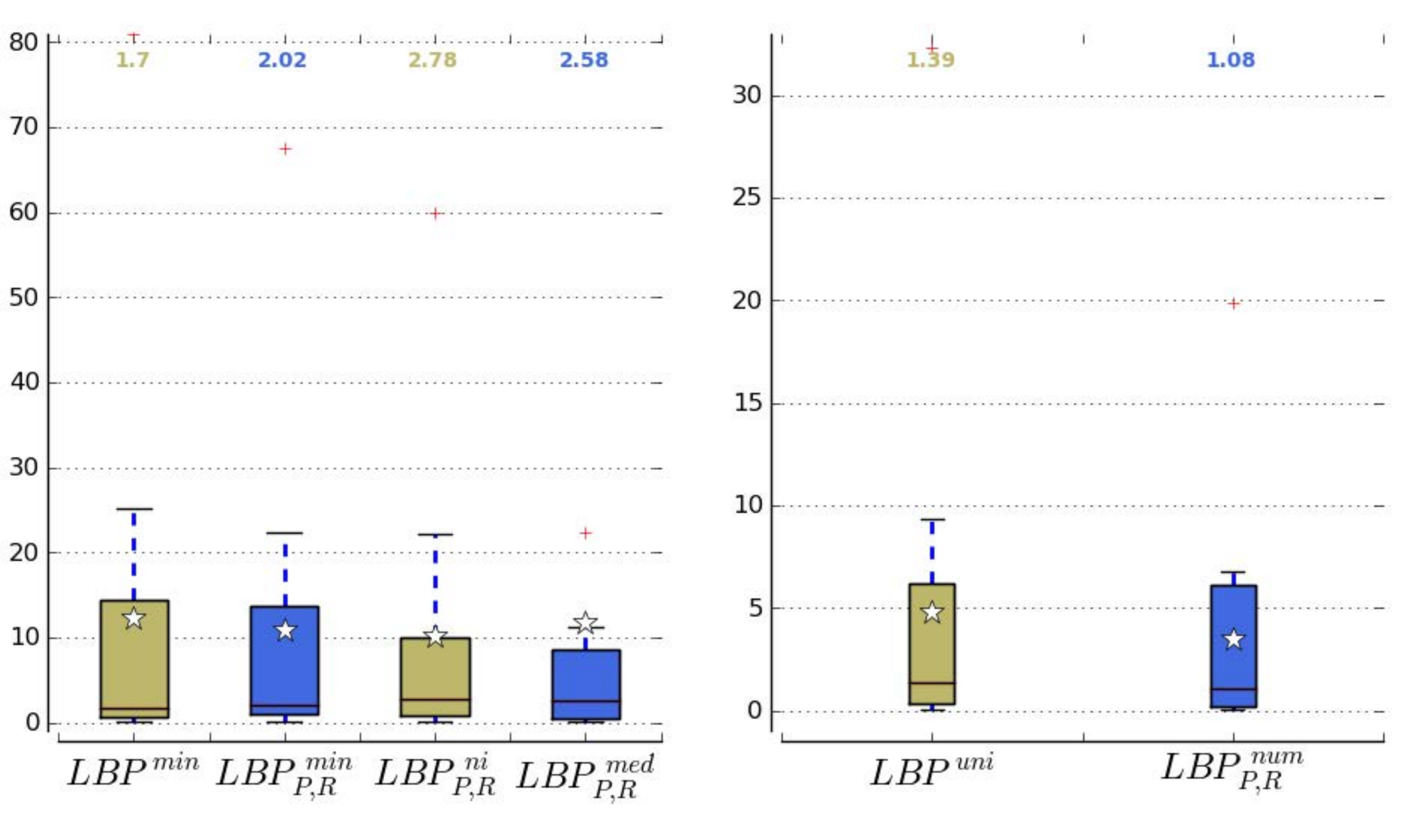}}
    \caption{Fig.~\ref{fig:fig_3a} mean distances using OD metric. Fig.~\ref{fig:fig_3b} mean distances using KL metric. }
    \label{fig:fig3} 
\end{figure}

We compared the performance of $LBP^{min}$, $LBP^{min}_{P,R}$, $LBP^{uni}_{P,R}$, $LBP^{num}_{P,R}$, $LBP^{ni}_{P,R}$, and $LBP^{med}_{P,R}$ in terms of accuracy using $P=8$ and $R=1$ on a circular neighborhood, (see Fig.~\ref{fig:fig3}). This configuration has shown good results in discriminating similar textures. We computed a set of OD and KLD values by compared each reference image in the database with its rotated versions. Thus the most accurate technique is the one that brings the smallest mean. We followed the assessment methodology proposed in \cite{ORJUELA2011} by quantifying the distance between distinctive textures using both OD and KL metrics, Fig.~\ref{fig:fig_3a} and Fig.~\ref{fig:fig_3b} respectively. 

Since the seven LBP approaches are rotational invariant is expected that distances are to be zero or close to zero but due to the fact that the rotated textures were scanned using a $512\times 512$ pixel video digitizing camera, the CCDs may has introduced some values that produce higher distances among LBP histograms.

%-------------------------------------------------------------------------
\subsection{Neighborhood size~\label{sec:size}}
An important issue of original LBP is the neighborhood size. It has small spatial support area, hence the bit-wise therein made between two single intensity value pixels is affected by noise. The next experiment was aimed to assess the radius size influence in texture classification. We present the classification performance of five LBP approaches with different radius $R = \left\{1,2,3\right\}$.

\begin{figure}[htbp]
    \centering
    \subfigure[]{\label{fig:fig_4a}\includegraphics[width=0.49\textwidth]{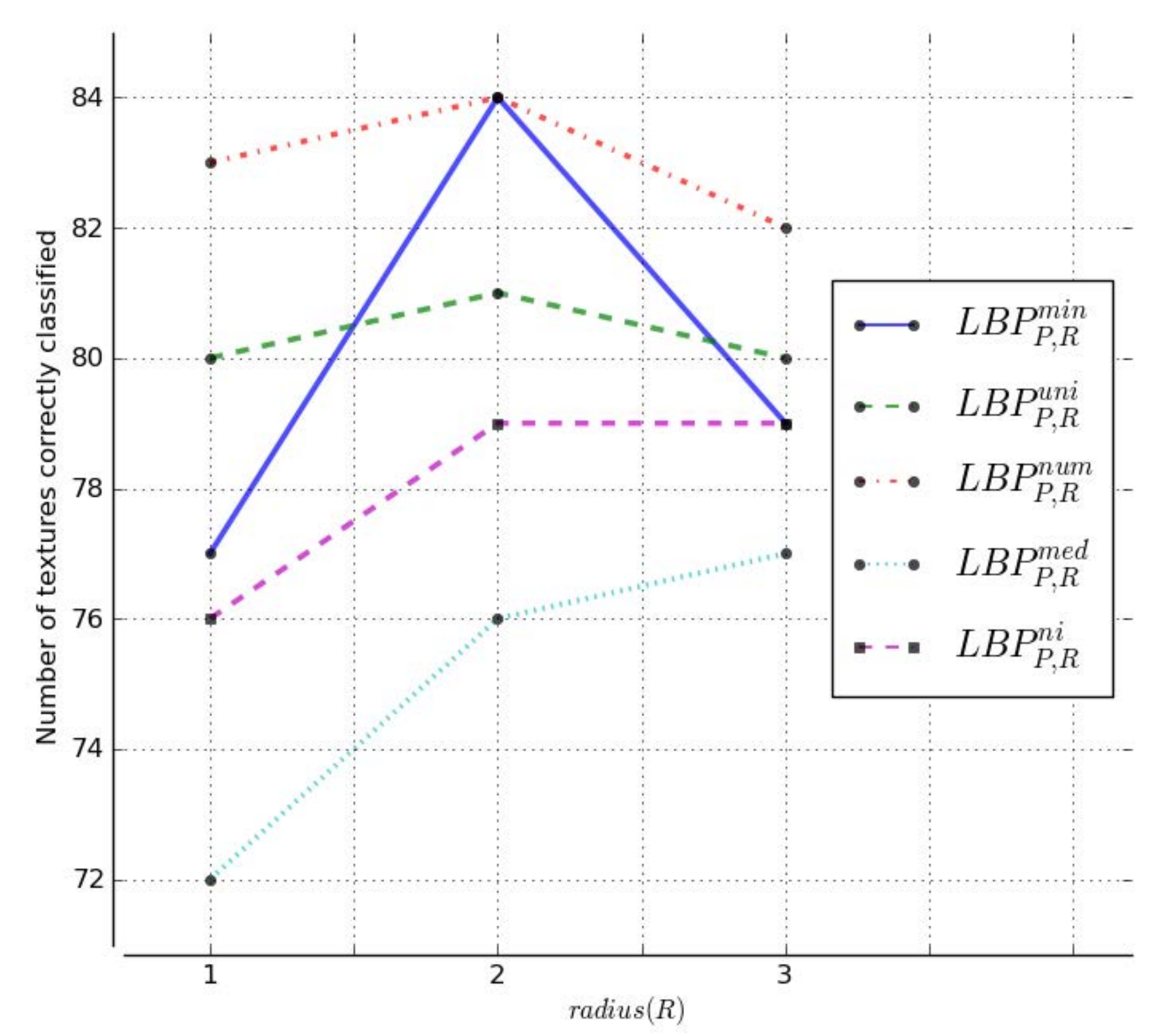}}
    \subfigure[]{\label{fig:fig_4b}\includegraphics[width=0.49\textwidth]{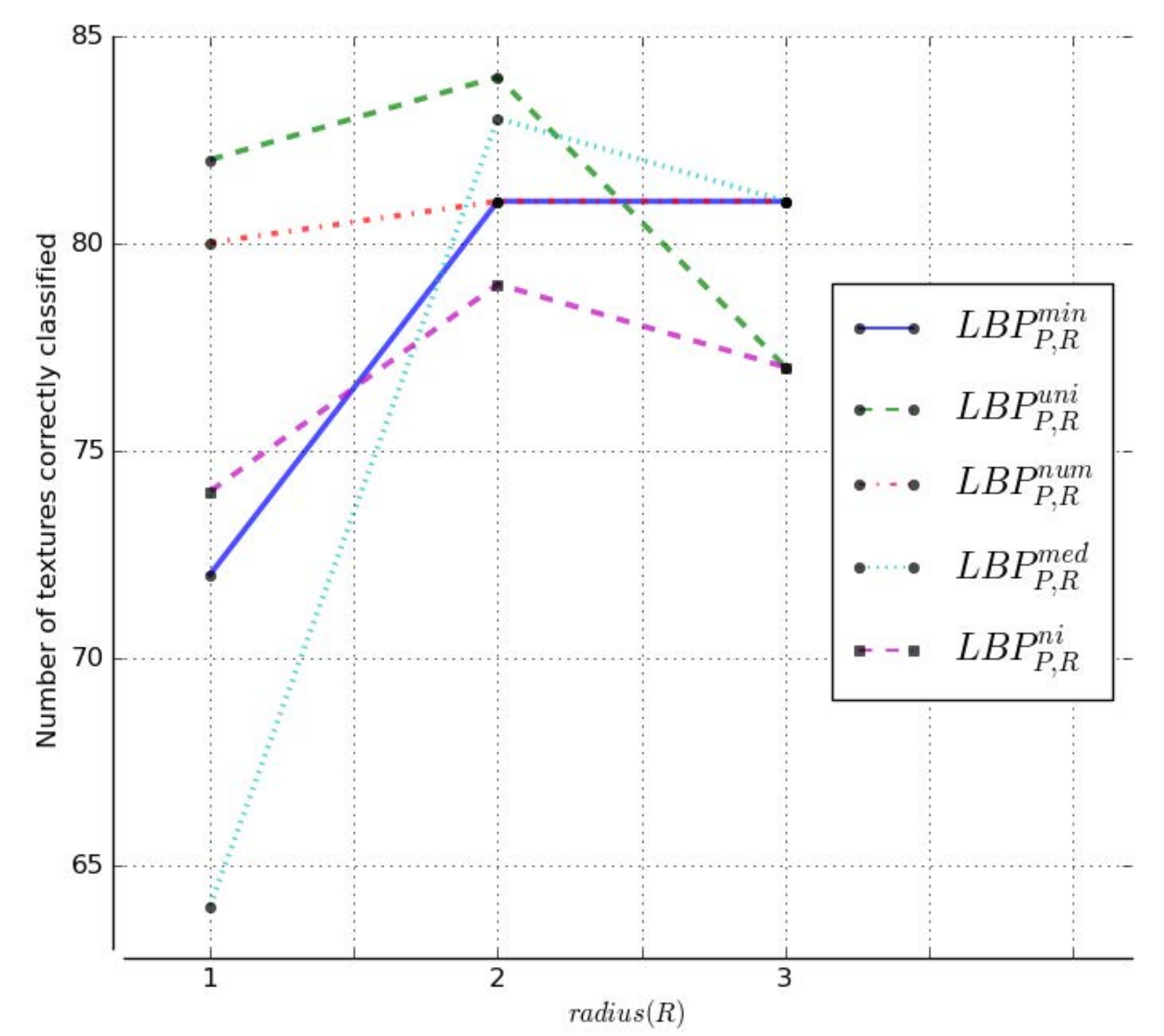}}
    \caption{AR of the LBP approaches using different radius sizes $R = \left\{1,2,3\right\}$. Fig.~\ref{fig:fig_4a} OD metric. Fig.~\ref{fig:fig_4b} KL metric.}
    \label{fig:fig4} 
\end{figure}

Fig.~\ref{fig:fig4} shows the classification performance comparison among five LBPs. In all cases the highest classification rate was achieved with $R=2$. On the contrary, the increased size of the radius caused a poor classification rate starting from $R=3$.

%-------------------------------------------------------------------------
\subsection{Noise~\label{sec:noise}}
\begin{figure}[htbp]
    \centering
    \includegraphics[width=0.5\textwidth]{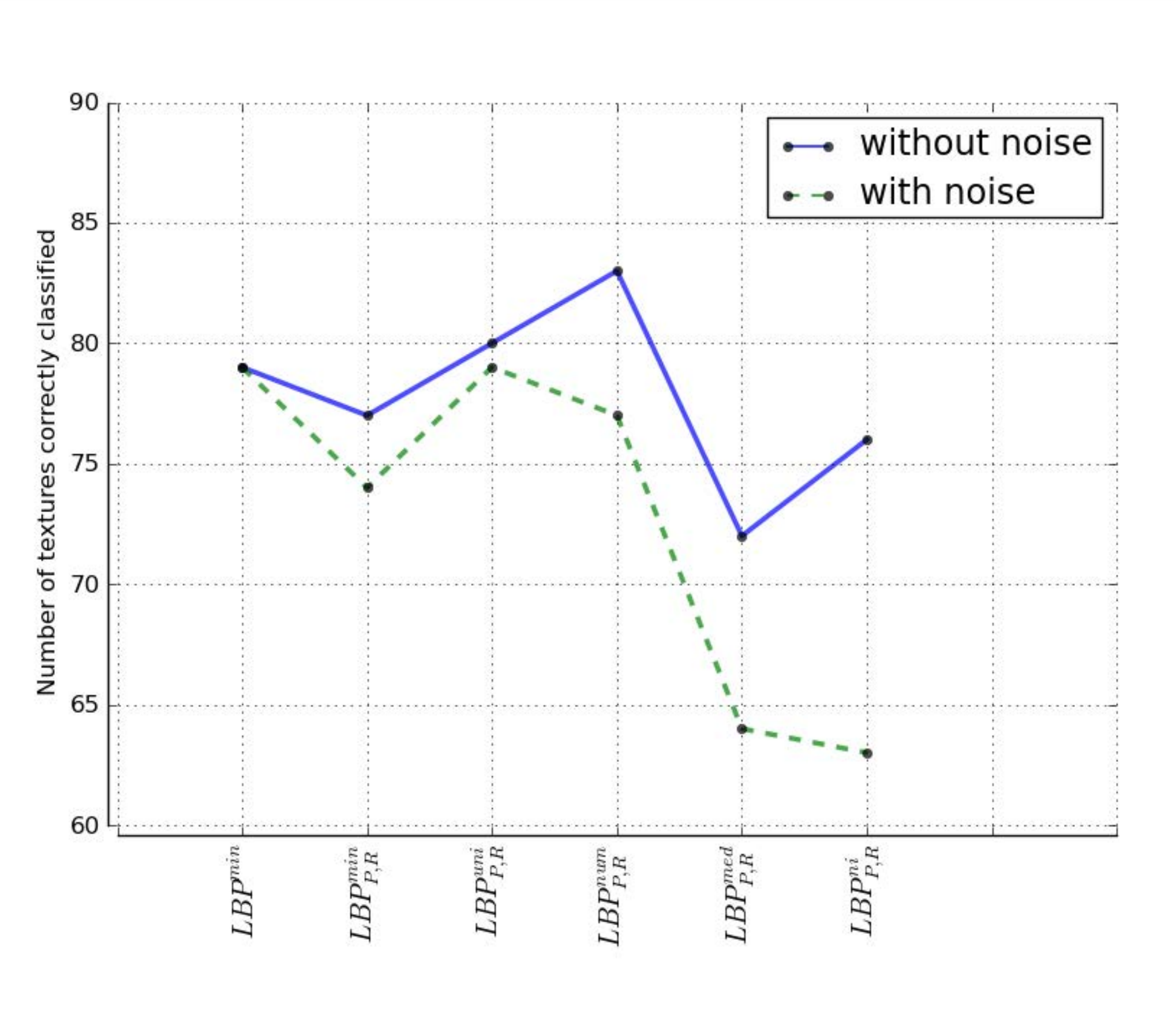}
    \caption{Performance of LBP approaches under additive Gaussian noise with media $\mu = 0$ and $\sigma{^2} = 0.06$. }
    \label{fig:fig_9} 
\end{figure}

LBP approaches are very sensitive to noise specially when a small neighborhood  is used. Since the amount of information associated to a pixel is not very large, even a small  pixel value change due to noise could lead to a different LBP label. Fig.~\ref{fig:fig_9} presents the accuracy rate of LBPs under the influence of additive Gaussian noise using OD metric. In this experiment we used the USC-SIPI database described in the preceding section. Gaussian noise with $\mu = 0$ and $\sigma^{2} = 0.06$ was added to the image database. This addition was implemented using Matlab \textit{imnoise} function.

Since Gaussian noise affects the AR, LBP demands to use preprocessing denoising step or normalization stage in order to avoid noise artifact influence.

%-------------------------------------------------------------------------
\subsection{Illumination~\label{sec:illumination}}
Lighting variation is one of the major challenge for the current feature descriptors. \cite{TAN2010b} presented a study of texture analysis under difficult lighting conditions and claims that LBP performance decreases almost exponentially under extreme illumination conditions. LBP by itself is not invariant to illumination changes and does not address the contrast of textures which is important in the discrimination. For this purpose, we are interested in combined LBP operators with a contrast information measure (CI), Eq.~(\ref{eq:eq18}). However, CI produces continuous values which need to be quantized. \cite{OJALA1994} proposed to quantize contrast values so that all bins have an equal number of elements. So far, setting the number of bins is still an open issue.

LBP and CI histograms could be combined in two ways: jointly or hybridly, \cite{GUO2010a}. In the first one, similar to 2D joint histograms, we can build a 3D joint histogram of them. In the second way, a large histogram is built by concatenating both LBP and CI histograms to form the so-called ``pseudo joint histogram''.

\begin{equation}
    CI = \sum^{m-1}_{i=0}{G_{i}}- \sum^{n-1}_{i=0}{g_{i}}
    \label{eq:eq18}
\end{equation}
where $G_{i}$ are the $m$ neighbor pixels greater or equals than $g_{c}$ and $g_{i}$ are the $n$ neighbor pixels lesser that $g_{c}$. 

In the next experiment we performed classification of textures under different illumination conditions using the CUReT database.

\begin{figure}[htbp]
    \centering
    \subfigure[]{\label{fig:fig_5a}\includegraphics[width=0.49\textwidth]{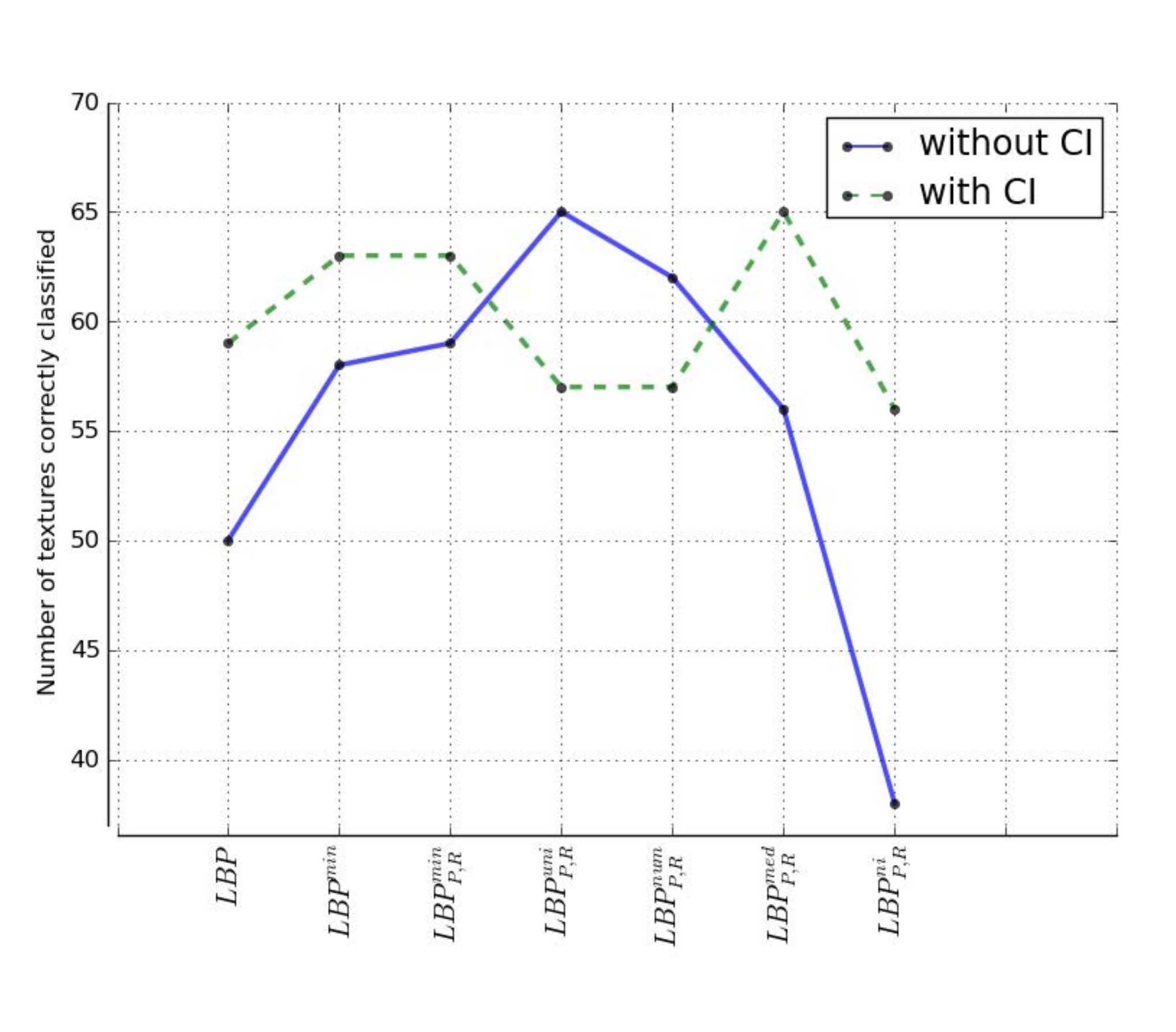}}
    \subfigure[]{\label{fig:fig_5b}\includegraphics[width=0.49\textwidth]{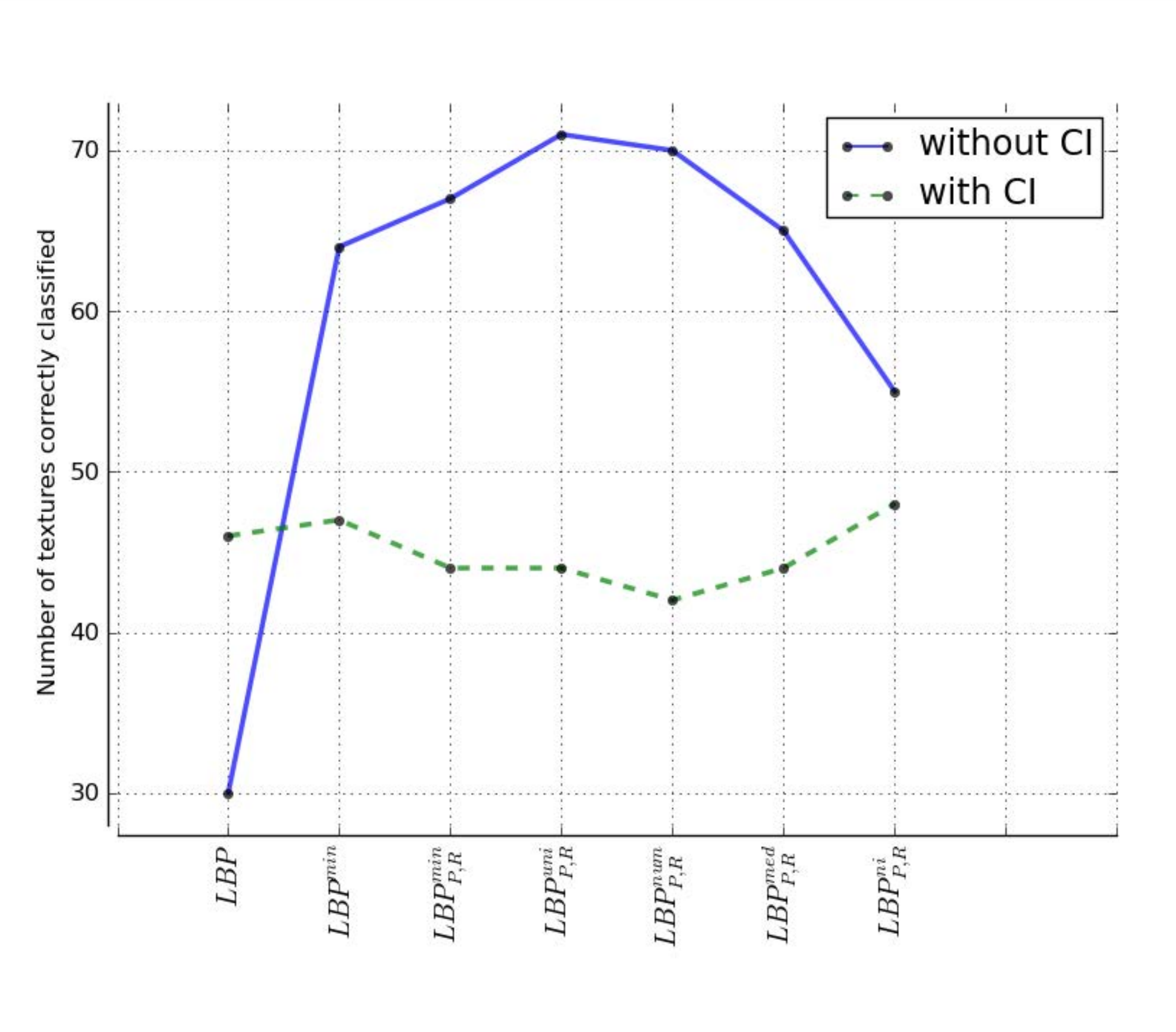}}
    \caption{Texture classification under illumination changes. Fig.~\ref{fig:fig_5a} shows performance of seven LPB/CI approaches using the OD metric. Fig.~\ref{fig:fig_5b}  shows performance of seven LPB/CI approaches using the KL metric}
    \label{fig:fig5} 
\end{figure}

As we expected, the AR achieved are lower that the results presented for rotational invariant analysis because LBP methodology is not illumination invariant. Nevertheless, CI improved the AR of almost every LBP methods using OD metric except for those methods based on uniform patterns, where their LBP histogram contains $P+1$ bins. In these cases, we need to apply other methodologies that are out of the scope of this study. On the contrary, in Fig.~\ref{fig:fig_5b} CI decreases the classification performance. A possible explanation is that CI entropy influences negatively KL metric leading a misclassification.

The combination of LBP and CI performs well in the case of rotational cases too. For rotational invariant experiments, the AR of original $LBP$ using the USC-SIPI image database are $38.46\%$ and $42.86\%$ for OD and KL metrics respectively. However, the AR increased up to $76.92\%$ and $78.02\%$ for OD and KL metrics respectively by adding CI histograms.

%-------------------------------------------------------------------------
\section{Conclusions~\label{sec:conclusions}}
LBP descriptors have been powerful tools for feature encoding. They have been successfully used in many different image analysis applications, in particular in the area of face recognition due to their excellent properties and computational simplicity. Since original LBP proposal has many limitations such as noise sensibility and it is affected by rotational transforms, a large number of extensions have been proposed. We have presented a LBP state of art for invariant and non-invariant rotational approaches. We evaluated the performance of several LBP algorithms proposed in the literature for texture classification. The results can be summarized into three groups: \begin{inparaenum}[\itshape i\upshape)]
\item LBPs based on minimal chains: $LBP^{min}$ and $LBP^{min}_{P,R}$ compute a minimal chain. Their performance could be affected by noise because if a pixel intensity value is disturbed the final LBP label changes.
\item LBPs based on neighborhood values: Prior a LBP label computation, $LBP^{ni}_{P,R}$ and $LBP^{med}_{P,R}$ perform a weighted average of neighboring pixels to minimize the effects of noise.
\item LBPs based on uniform values: $LBP^{uni}_{P,R}$, $LBP^{num}_{P,R}$ compute a uniformity measure prior LBP label computation which corresponds to the number of spatial transitions in the pattern. Since $LBP^{num}_{P,R}$ add extra information of non-uniform patterns into the LBP histogram it has a better texture classification performance than $LBP^{uni}_{P,R}$.
\end{inparaenum} 
Further work includes extending this techniques by applying a preprocessing stage based on Gabor filtering for increasing the robustness to illumination and noise degradations. Another extension will be based on the use of FPGAs to reduce the computational time.

%-------------------------------------------------------------------------
\section*{Acknowledgments}
This work has been partially supported by the following UNAM grants: PAPIIT~IN113611 and IXTLI~IX100610 and by the Spanish Ministry of Science and Technology under projects TEC2010-20307 and TEC2010-09834-E.

%-------------------------------------------------------------------------
\bibliographystyle{model2-names}
\bibliography{references}

\end{document}